\documentclass{article}
\usepackage{iclr2026_conference,times}

\usepackage[utf8]{inputenc}
\usepackage[T1]{fontenc}
\usepackage{microtype}
\usepackage{amsmath,amssymb,amsfonts,amsthm}
\usepackage{mathtools}
\usepackage{bm}
\usepackage{graphicx}
\graphicspath{{../}}
\usepackage{booktabs}
\usepackage{multirow}
\usepackage{xcolor}
\usepackage{hyperref}
\usepackage{url}
\usepackage{cleveref}
\usepackage{algorithm}
\usepackage{algpseudocode}
\usepackage{enumitem}

\usepackage{amsmath,amsfonts,bm}

\def\eqref#1{equation~\ref{#1}}

\def\1{\bm{1}}

\def\mM{{\bm{M}}}

\DeclareMathAlphabet{\mathsfit}{\encodingdefault}{\sfdefault}{m}{sl}
\SetMathAlphabet{\mathsfit}{bold}{\encodingdefault}{\sfdefault}{bx}{n}

\newcommand{\E}{\mathbb{E}}

\newcommand{\R}{\mathbb{R}}

\iclrfinalcopy   

\renewcommand{\R}{\mathbb{R}}
\renewcommand{\E}{\mathbb{E}}

\newcommand{\activations}{\mathcal{A}}
\newcommand{\behavior}{\mathcal{Y}}
\newcommand{\codomain}{\mathcal{C}}
\newcommand{\concepts}{\mathcal{Z}}
\newcommand{\manifold}{\mathcal{M}}
\newcommand{\Mh}{\mathcal{M}_h}
\newcommand{\My}{\mathcal{M}_y}

\newcommand{\encoder}{\phi}
\newcommand{\decoder}{\psi}
\newcommand{\Jenc}{J_{\encoder}}

\newcommand{\pullback}{g}

\newcommand{\activation}{h}

\newcommand{\dir}[1]{\mM{\rm d}{#1}}
\newcommand{\dgen}{\ensuremath{\mathbf{\tt d}}}

\newcommand{\bhattacharyya}{\dgen_{\mathrm{BC}}}
\newcommand{\hellinger}{\dgen_{H}}

\newcommand{\widthclamp}[1]{%
  \resizebox{\ifdim\width>\linewidth\linewidth\else\width\fi}{!}{#1}}

\title{Riemannian-Manifold Steering:\\
       Geometry-Aware Generative Autoencoders for Label-Free Steering}

\author{Narmeen Oozeer \\
Martian
\And
Shivam Raval \\
Harvard University
\And
Philip Quirke \\
Martian
\And
Manikandan Ravikiran \\
Thoughtworks
\And
Jeff Phillips \\
University of Utah
\And
Shriyash Upadhyay \\
Martian
\And
Amirali Abdullah \\
Thoughtworks
}

\date{}

\begin{document}
\maketitle

\begin{abstract}

Steering a language model - intervening on its internal activations to change downstream behaviour - has recently expanded beyond linear interpolation to nonlinear methods such as angular and kernelized steering, which define intervention transformations without learning an explicit geometry over paths in activation space. Freshly introduced geometry-aware manifold methods do learn such a geometry, but require labelled class centroids together with prescribed cyclic or sequential structure. These assumptions restrict where manifold steering can be applied, since existing constructions require labelled centroids and compatible boundary conditions. We recast manifold steering more broadly as \textbf{Riemannian geodesic computation} on activation space, recovering linear and labelled-spline steering as geodesics under particular choices of metric. A principled metric within this framework is the output-space Hellinger distance pulled back to activations; we approximate this with a learned encoder trained on output distances over a small concept-token schema - no per-prompt labels, no topology prior, and no per-task curve fitting. Empirically, the method reliably drives the model onto the target class across all tasks in a standard four-task language-model arithmetic benchmark, while following more behaviourally natural trajectories than baselines on smaller output spaces. We thereby provide a unified Riemannian framework for manifold steering together with a schema-supervised, label-free instantiation that operates without labelled centroids or prescribed boundary conditions.

\end{abstract}

\section{Introduction}
\label{sec:intro}

\begin{figure}[!ht]
\centering
\IfFileExists{figures/gaga_steering_diagram.png}
  {\includegraphics[width=0.8\linewidth]{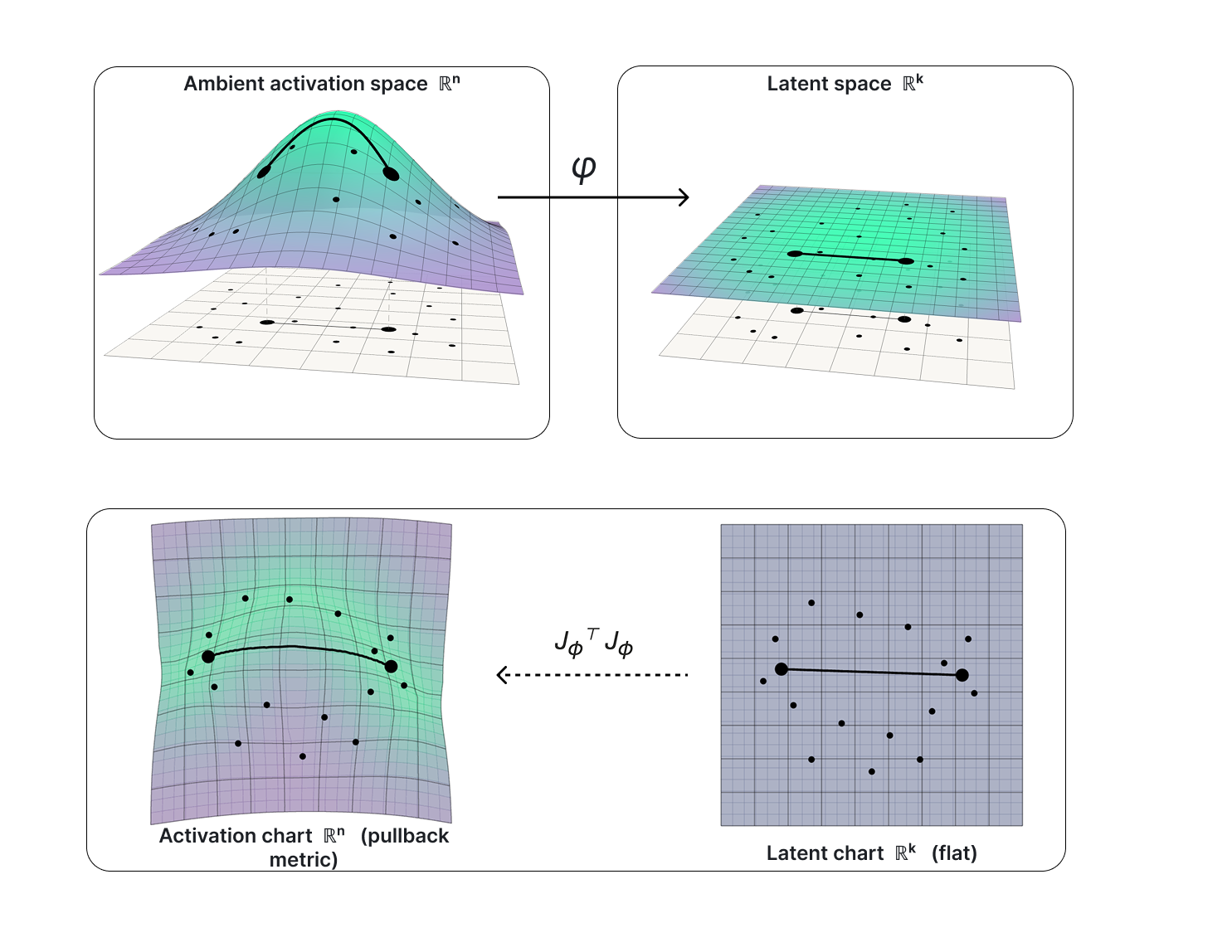}}
  {\fbox{\parbox{0.92\linewidth}{\centering\small\vspace{2em}
   \textbf{[figure pending]}\\[0.5em]
   Export the \texttt{GAGA-steering} frame from Figma (file
   \texttt{G6P0GevKauWYMVQLoZ7ILY}) and save it to
   \texttt{paper/figures/gaga\_steering\_diagram.png}.\vspace{2em}}}}
\caption{\textbf{Manifold steering as Riemannian geodesic computation.}
A GAGA encoder $\varphi$ maps the ambient activation space
$\activations = \R^n$, a curved activation manifold (top-left),
into a latent space $\R^k$ (top-right). The encoder is trained so that
latent Euclidean distance reproduces a target distance $\dgen$
between training points: either the PHATE diffusion distance on
activations (activation geometry) or the Hellinger distance
$\hellinger$ on output distributions (behaviour geometry). Either choice
is represented within the same flat latent chart: $\varphi$ learns a
coordinate system on $\R^k$ in which the chosen target distance becomes
ordinary Euclidean distance, so geodesics are straight lines there
(bottom-right). The geometry resurfaces when this flat metric is pulled
back through the encoder Jacobian
$J_\varphi^{!\top} J_\varphi$, warping the activation chart
(bottom-left) so steering trajectories curve to follow the learned
geometry.}
\label{fig:gaga-diagram}
\end{figure}

Steering a neural network, intervening on internal activations to
controllably change downstream behaviour, benefits from respecting
the curved, low-dimensional structure that recent work argues
language-model activations exhibit
\citep{wurgaft2026manifold, vu2026angular, raval2026curveballsteeringrightdirection,
engels2024notall, modell2025, park2025representations, kantamneni2025helix, 
gurnee2024, karkada2026origins, prieto2026}. We do not claim a true
smooth low-dimensional activation manifold exists in any model-theoretic
sense: the data we operate on are a few-hundred distinct facts
$\times{\sim}21$ paraphrases per task, finite and clustered. The claim
we work from, and that prior work supports empirically, is weaker: a
useful low-dimensional \emph{approximation} can be learned, and
interventions that follow it produce smoother behavioural trajectories
than interventions that ignore it. Interventions that ignore the
structure altogether (a straight line in activation space) either fail
to move the model or produce incoherent intermediate states; the
interventions that follow it (a fitted spline) produce smooth
behavioural trajectories but at a price in supervision.

\citet{wurgaft2026manifold} make this concrete on language-model
activations: they fit a cubic spline through per-class activation
centroids and steer along its arc. Their construction works, but it
requires the analyst to supply (i) per-class activation centroids as
spline knots and (ii) a knot ordering with a boundary-condition choice
(periodic for weekdays and months, natural for letters and ages) that
commits to the conceptual topology before fitting. The class labels
and the boundary-condition choice both do real work; relax either and
the construction does not apply.

\citet[\S3.4]{wurgaft2026manifold} take a first step toward unifying
steering strategies geometrically, defining three analytical metrics
($G_I$ flat, $G_E$ density-based, $G_F$ output pullback) under which
their linear, manifold, and pullback steering arise as geodesics. We
approach the same problem from a different, more data-driven angle: rather than
specifying the metric analytically, we \emph{learn} it from data via
an encoder pullback (\cref{fig:gaga-diagram}), which (i) makes metric
and solver orthogonal design axes and (ii) opens supervision modes
(schema-supervised output distances, in our case) that analytical
metrics cannot use. It also reveals that the existing alternatives are not really alternatives. Linear interpolation is the geodesic under $g = I$. \citet{wurgaft2026manifold}'s labelled cubic spline arises in their framework as the geodesic of a density-based metric $G_E$ whose minima coincide with the spline
through the class centroids; restricted to that spline, $G_E$-geodesics agree with the first fundamental form $g = J_s^{\!\top} J_s$ of the spline parameterisation $s$. Both baselines are then special cases of our framework, not competing methods.

The principled choice of $g$ within this framework is the pullback of
the output-space Hellinger metric through the rest-of-the-network
forward map $F$ \citep{wurgaft2026manifold, amari2016information}:
$G_F = J_F^{\!\top} g_y J_F$. This is the metric \citet{wurgaft2026manifold}
formally specify but do not realize via explicit-Jacobian autodiff:
they access $G_F$ implicitly by directly optimizing activation paths to
induce a target trajectory on $\My$ (their \S3.3, $R^2 \in [0.47,
0.78]$). Computing $J_F$ explicitly requires per-pair passes through
$F$ and is expensive as a training signal at scale. We replace this analytical metric with a
cheap learned \emph{surrogate}: a Geometry-Aware Generative Autoencoder
\citep[GAGA;][]{sun2025gaga} whose encoder $\varphi$ is trained to
match a target distance source, and whose induced pullback
$g_\varphi = J_\varphi^{\!\top} J_\varphi$ acts as a stand-in for
$G_F$. The surrogate is not equal to $G_F$ (different supervision
sources target different geometries, and the encoder's geometry can
diverge from $F$'s along the path), but we evaluate it empirically
on the same behavioural metric ($E_{\mathrm{BC}}$ on output
distributions) that $G_F$ would optimise. Crucially, the encoder is
supervised by an output-space distance over a small fixed
\emph{concept-token schema}: it requires no per-prompt class labels
and no boundary-condition prior. We refer to this as \emph{schema-supervised,
label-free}, not unsupervised, because the choice of concept
tokens (e.g.\ the seven weekday tokens) and the evaluator that scores
output Hellinger distances are fixed inputs to the training pipeline,
even though no individual prompt is labelled.

\paragraph{Contributions.}
\begin{enumerate}
\item \textbf{Framework.} \citet[\S3.4]{wurgaft2026manifold} cast
      linear, manifold, and pullback steering as geodesics under three
      analytically specified metrics ($G_I$, $G_E$, $G_F$). We reframe
      metric choice as a \emph{learning} problem: rather than designing
      $G$ from a density or energy prior, we learn an encoder $\varphi$
      whose pullback $J_\varphi^{\!\top} J_\varphi$ approximates the
      principled target $G_F$ from data. This change of lens (i)
      separates metric from solver as orthogonal design axes (letting
      any metric pair with any compatible solver) and (ii) admits
      supervision sources unavailable to analytical metrics, including
      the output-Hellinger distances we use, which require no labelled
      centroids and no boundary-condition prior (\cref{sec:framework}).
\item \textbf{Method.} We propose a cheap, schema-supervised
      \emph{surrogate} for $G_F$: a GAGA encoder trained on
      output-Hellinger distances over a small concept-token schema,
      label-free at the prompt level and boundary-condition-free
      (\cref{sec:gaga}).
\item \textbf{Empirical Results.} On the four arithmetic tasks of
      \citet{wurgaft2026manifold} the output-grounded raw-$4096$
      variant lands the target class decisively on every task
      (waypoint legibility $\geq 99\%$; endpoint target probability
      $1.4$--$5.7\times$ the better baseline), wins behaviour-fidelity
      $E_{\mathrm{BC}}$ on the two small output spaces (weekdays $7$:
      $12\times$; months $12$: $2.4\times$), and loses
      $E_{\mathrm{BC}}$ on the two larger ones (letters $22$, ages
      $91$); a 22-cell PCA($64$) encoder-variant ablation shows the
      effect is specific to the raw-$4096$ output-grounded
      representation (\cref{sec:results}).
\end{enumerate}

A reader new to Riemannian-geometry terminology may find \cref{sec:glossary}
useful; a reader who prefers to skip the framing and read empirics first may
go directly to \cref{sec:gaga}.

\section{Notation and glossary}
\label{sec:glossary}

This section assembles terminology used throughout the paper, oriented toward
readers from mechanistic interpretability who may be encountering
manifold-learning concepts for the first time. We assume comfort with linear
algebra, basic probability, and standard transformer activations. Definitions
here are deliberately informal; pointers to standard textbook treatments
\citep{lee2003smooth} are given where appropriate.

\paragraph{Manifold (smooth, embedded).}
A $d$-dimensional smooth manifold $\mathcal{M} \subset \R^n$ is a subset that
locally resembles $\R^d$: every point has a neighborhood that can be smoothly
mapped to and from an open subset of $\R^d$. Throughout this paper $d \ll n$:
activation manifolds are typically 1- or 2-dimensional structures embedded in
$n=4096$-dimensional residual streams.

\paragraph{Intrinsic vs.\ ambient dimension.}
The \emph{ambient} dimension is the dimension of the space the manifold lives
in ($n$); the \emph{intrinsic} dimension is the dimension of the manifold
itself ($d$). For days of the week, the ambient dimension is the
residual-stream width; the intrinsic dimension is one (a circle).

\paragraph{Riemannian metric / metric tensor.}
A Riemannian metric $g$ assigns to each point a positive (semi-)definite
bilinear form $g(\activation)$ on the tangent space at $\activation$. In
coordinates, $g(\activation)$ is an $n \times n$ symmetric positive
(semi-)definite matrix. The metric defines length:
$L_g(\pi) = \int_0^1 \sqrt{\dot\pi(t)^{\!\top} g(\pi(t)) \dot\pi(t)}\, \dir t$.
Different metrics produce different notions of ``shortest path''.

\paragraph{Geodesic.}
A geodesic between $\activation_0$ and $\activation_1$ under metric $g$ is the
path that minimizes $L_g$. Under the flat metric $g = I$, geodesics are
straight lines; under non-flat metrics they curve to follow the manifold's
intrinsic geometry.

\paragraph{Pullback metric.}
Given a smooth map $\encoder : \activations \to \R^k$ (e.g.\ an encoder), the
\emph{pullback} of the Euclidean metric on $\R^k$ through $\encoder$ is the
metric
\[
\pullback(\activation) \;=\; \Jenc(\activation)^{\!\top}\Jenc(\activation),
\qquad
\Jenc(\activation) \;=\; \frac{\partial \encoder}{\partial \activation}(\activation) \in \R^{k \times n}.
\]
This metric measures distances in $\activations$ as they ``feel'' from
$\encoder$'s perspective: two points are far apart under $\pullback$ if their
encodings are far apart in $\R^k$. Pullback metrics transport geometric
structure from a chosen latent space back to the original activation space.

\paragraph{Encoder Jacobian and $\Jenc^{\!\top}\Jenc$.}
For a trained encoder, $\Jenc^{\!\top}\Jenc$ encodes how much each ambient
direction stretches or compresses under $\encoder$. Directions \emph{on} the
activation manifold are mapped near-isometrically (eigenvalues
$\approx 1$); directions \emph{off} the manifold are heavily compressed
(small or zero eigenvalues). This is what gives the pullback metric its
manifold-respecting character: moving on-manifold is cheap, moving
off-manifold is expensive.

\paragraph{Isometry / scaled isometry.}
Two metric spaces $(\mathcal{M}_1, \dgen_1)$ and $(\mathcal{M}_2, \dgen_2)$ are
\emph{isometric} if there is a bijection $\phi$ with
$\dgen_2(\phi(x), \phi(y)) = d_1(x, y)$ for all $x, y$; \emph{scaled isometric} if
$\dgen_2(\phi(x), \phi(y)) = c \cdot \dgen_1(x, y)$ for some $c > 0$.

\paragraph{Activation space $\activations$.}
$\activations = \R^n$, the residual stream at a chosen (layer, token-position)
of the transformer. Throughout we use Llama-3.1-8B layer 28, so $n = 4096$.

\paragraph{Probability simplex $\Delta^k$.}
$\Delta^k = \{p \in \R^{k}_{>0} : \sum_i p_i = 1\}$: $k$-class probability
distributions with strictly positive entries. Restricting to strictly positive
entries makes $\Delta^k$ a smooth manifold of dimension $k-1$.

\paragraph{Behavior space $\behavior$.}
$\behavior = \Delta^{|\concepts|}$, the open probability simplex over a
conceptual domain $\concepts$. A point in $\behavior$ is a distribution over
concept tokens (e.g.\ over the seven days of the week, plus an ``other''
class). We embed $\behavior$ in Hellinger space via $p \mapsto \sqrt{p}$ to
linearize.

\paragraph{Hellinger distance.}
$\hellinger(p, q) = \tfrac{1}{\sqrt{2}}\,\|\sqrt{p} - \sqrt{q}\|_2$. The
embedding $p \mapsto \sqrt{p}$ takes $\Delta^k$ to the positive orthant of the
unit sphere in $\R^k$; under this embedding, Hellinger distance is just
Euclidean distance.

\paragraph{Bhattacharyya distance.}
$\bhattacharyya(p, q) = -\log \sum_i \sqrt{p_i q_i}
= -\log\langle\sqrt{p}, \sqrt{q}\rangle$. We use it to measure naturalness of
a behavioral trajectory: low values mean the trajectory stays close to the
manifold of natural model outputs.

\paragraph{Linear representation hypothesis (LRH).}
The hypothesis that high-level features are encoded as linear functions of
activations \citep{park2023linear, elhage2022superposition}. Under LRH,
steering reduces to vector arithmetic. Empirical evidence is mixed: LRH holds
for some features but not for others, including the cyclic and graph-structured
concepts we study.

\paragraph{Activation centroid.}
The mean of activations sharing a label (e.g.\ all activations whose target
answer is ``Wednesday''). We use centroids only at evaluation time;
\citet{wurgaft2026manifold} additionally use them at fit time, as the knots of
their cubic-spline activation manifold.

\paragraph{PHATE diffusion-potential distance.}
PHATE (Potential of Heat-diffusion for Affinity-based Transition Embedding) \citep{moon2019phate} computes pairwise distances by (a) building a
$k$NN graph over the data, (b) running a diffusion process on it for $t$
steps, and (c) measuring a log-difference of resulting distributions. PHATE
distances respect manifold geodesics: two points close in Euclidean distance
but on different parts of the manifold get a large PHATE distance. GAGA's
distance-matching loss is supervised by PHATE distances.

\paragraph{Steering / activation steering.}
Modifying internal activations during a forward pass to alter behavior. The
classical recipe is \emph{linear steering}, $h \mapsto h + \alpha v$.
\emph{Manifold steering} replaces this with paths that follow the activation
manifold's curvature. \emph{Learned manifold steering} (this work) replaces it
with paths that follow a manifold recovered from the data.

\paragraph{Causal mediation / intervention.}
The methodology of replacing a chosen activation with a target value during a
forward pass and observing the effect on output \citep{geiger2021causal,
geiger2025causal}. We use \texttt{pyvene}-style interventions
\citep{wu2024pyvene} as a black box.

\paragraph{Naturalness energy $E_{\mathrm{BC}}$.}
For a behavioral trajectory $\gamma : [0,1] \to \behavior$ induced by a
steering path,
\(
E_{\mathrm{BC}}(\gamma) = \int_0^1 \bhattacharyya(\gamma(t), \My)\, \dir t
\)
\citep{wurgaft2026manifold}. Low $E_{\mathrm{BC}}$ means the trajectory stays in
the region of plausible model outputs.

\paragraph{$\Mh$, $\My$.}
$\Mh \subset \activations$ is the activation manifold (this paper: the image
of GAGA's decoder). $\My \subset \behavior$ is the behavior manifold (a cubic
spline through Hellinger-embedded behavior centroids; we keep this identical
to \citet{wurgaft2026manifold}, see \cref{sec:experiments}).

\section{Manifold steering as Riemannian geodesic computation}
\label{sec:framework}

Given a Riemannian metric $g$ on activation space $\activations = \R^n$,
steering between two endpoints $\activation_0, \activation_1$ is the
geodesic of $g$ joining them. This is the framing this paper adopts. Under
it, the design space for a steering method has two orthogonal axes:
\emph{which metric} we put on $\activations$, and \emph{how we compute
geodesics} under that metric. The first is a modeling choice about the
geometry that the LM induces on its own activations; the second is a
numerical-methods choice. \citet[\S3.4]{wurgaft2026manifold} name three analytical metrics ($G_I$,
$G_E$, $G_F$) and tie each to a single solver (closed-form for linear and
spline, path-level L-BFGS for pullback). We instead treat the metric itself
as something to be learned from data, and metric and solver as orthogonal
design choices. The learned-metric lens locates linear and labelled-spline
steering as analytical-metric special cases, identifies $G_F$ as the
principled analytical target, and admits a new class of methods
(learned-Jacobian surrogates) that the analytical-only framing does not.

The paper proceeds in three steps. In the remainder of this section we
motivate the framing itself: why a Riemannian formulation in particular
(\cref{sec:framework:why-riem}), and why the pullback construction is the
principled source of a metric within it (\cref{sec:framework:why-pullback}).
\Cref{sec:metrics} taxonomizes the four metrics this framework instantiates
on language-model activations and shows that the linear and labeled-spline
baselines are degenerate cases of the same construction.
\Cref{sec:solvers} taxonomizes the solvers we use under each metric.

\subsection{Why Riemannian geometry?}
\label{sec:framework:why-riem}

\paragraph{Linear interpolation is the Riemannian case $g = I$.}
The straight-line path
$\pi_{\text{lin}}(t) = (1-t)\activation_0 + t\activation_1$ is the geodesic
of the Euclidean metric $g = I$ on $\R^n$. The ``no-geometry'' baseline that
underwrites most prior steering work is therefore not outside the Riemannian
framework: it is the flattest point.

\paragraph{Parametric-curve steering is Riemannian under the induced metric.}
For any smooth injective parameterization $s: \R^k \to \activations$ with
image $\manifold = s(\R^k)$, the differential $J_s = \partial s / \partial u$
induces a Riemannian metric on $\manifold$ via the first fundamental form
$g_\manifold = J_s^{\!\top} J_s$ \citep{docarmo1992riemannian, lee2018riemannian}.
The geodesics of $g_\manifold$, traced in the intrinsic coordinate $u$, are
exactly the curve family of $s$ itself. \citet{wurgaft2026manifold}'s labeled
cubic spline is the case where $s$ is a periodic or natural cubic spline
fit through labeled class centroids and $\manifold$ is the resulting 1-D
submanifold of $\activations$. Their construction is therefore not an
alternative to a Riemannian framework: it is a particular Riemannian metric,
namely the one induced by a hand-specified parameterization.

\paragraph{Riemannian geometry is the maximal smooth, local, length-based framework.}
Any steering method that produces smooth paths by minimizing a local cost
functional is expressible as geodesic flow under some metric
\citep{jost2017riemannian}. Methods that fall outside this framework do so
by relaxing smoothness (e.g.\ discrete optimal transport), determinism
(diffusion-based steering), or locality (global path-optimization with
non-local constraints). We adopt the Riemannian framing because: (i) it
subsumes the existing baselines as special cases, (ii) it cleanly separates
``which geometry'' from ``how we compute paths under it,'' and (iii) it
gives access to the mature machinery of Riemannian solvers
\citep{arvanitidis2018latent, sun2025gaga}.\footnote{A careful reader will
note that ``smooth local length-based'' is more precise than ``any smooth
path-finding method.'' The strict generalization beyond Riemannian to Finsler
geometry permits anisotropic but homogeneous length functionals; for the
length functionals we use, $\int \sqrt{\dot\pi^{\!\top} g \dot\pi}\, \dir t$
with symmetric positive-definite $g$, the Riemannian case is sufficient.}

\subsection{Why a pullback metric?}
\label{sec:framework:why-pullback}

The framework above leaves open which metric we put on $\activations$.
There is no canonical answer, but there is a canonical \emph{construction}:
the pullback metric.

\paragraph{The pullback construction reduces metric design to target-space choice.}
For any smooth map $\phi: \activations \to \codomain$ with a meaningful
metric $g_\codomain$ on the codomain, the pullback
$g(\activation) = J_\phi(\activation)^{\!\top}\,g_\codomain\,J_\phi(\activation)$
is the unique metric on $\activations$ making $\phi$ a local isometry
\emph{into} $\codomain$ \citep[Ch.~13]{lee2003smooth}. We emphasise
that the isometry is to the codomain only (a definitional property),
not to whatever ``true'' behaviour geometry the model induces in
practice.
\paragraph{Rank deficiency and the $\epsilon I$ regulariser.}
When $\phi$ maps to a codomain of dimension $k \ll \dim\activations$,
the pullback $J_\phi^{\!\top} J_\phi$ has rank at most $k$ and is
therefore rank-deficient PSD, not a Riemannian metric on the full
ambient space. In every operational use of $G_\phi$ in this paper
(\cref{sec:metrics,sec:solvers}), we replace it by the regularised
form $G_\phi + \epsilon I$ with $\epsilon > 0$ small. This makes the
metric a bona-fide, strongly anisotropic Riemannian metric: the $k$
directions captured by $J_\phi$ retain their target-pullback weights,
while the remaining $\dim\activations - k$ directions are
all\-but-isotropic at cost $\epsilon$. Without the $\epsilon I$ term
the construction is degenerate (sub-Riemannian); the regulariser is
not a numerical convenience but a structural part of the metric we
actually use. \emph{Both the analytical $G_F$ formally specified by
\citet[Def.~1 eq.~7]{wurgaft2026manifold} and the learned encoder
pullback we propose (\cref{sec:gaga}) inherit this regularisation.}
\paragraph{Why pull back from outputs.}
Choosing a metric on activations
reduces to choosing a target space $\codomain$ whose geometry we already
trust, plus the differentiable map $\phi$. This is a familiar move in
information geometry, where the Fisher metric is itself a pullback of $L^2$
through a parametric family of distributions \citep{amari2016information},
and in deep generative modeling, where decoder Jacobians pull back from a
latent space \citep{arvanitidis2018latent, shao2018riemannian,
kalatzis2020variational, chadebec2022data}.

\paragraph{Output behavior space is the natural target for activation steering.}
Activations matter only insofar as they shape outputs: two activations that
decode to the same output distribution are behaviorally equivalent
regardless of their Euclidean distance. A metric on activation space should
therefore identify behaviorally-equivalent activations as having distance
zero and separate behaviorally-distinct ones. The Euclidean metric $g = I$
fails this: it weights all $4096$ residual-stream directions equally
despite most being behaviorally inert. The pullback from output behavior
space succeeds by construction: the metric is zero in directions where the
output is unchanged. \citet{wurgaft2026manifold}'s Definition~1 names
three metrics: $G_I$ (flat), $G_E$ (density), and $G_F$ (output
pullback). The third, $G_F = J_F^{\!\top} g_y J_F + \epsilon I$, is the
principled target for activation-space steering and is the metric our
learned surrogate approximates. With $\phi = F$ the rest-of-the-network
map from activations to output distributions (in Hellinger coordinates,
so that $g_\codomain = I$), this is exactly
$G_F = J_F^{\!\top} J_F + \epsilon I$, the metric
\citet[Def.~1 eq.~7]{wurgaft2026manifold} formally specify and access
implicitly via path-level L-BFGS but do not realize via explicit
Jacobians. The framework's principled metric has a $25$-year information-geometric
pedigree.

The remaining design choice is how to instantiate the principled
output-pullback in practice. We examine three instantiations in
\cref{sec:metrics}: the flat case ($g = I$), the implicit case (the
spline-induced first fundamental form), and a learned-encoder
\emph{surrogate} (GAGA's $J_\varphi^{\!\top} J_\varphi + \epsilon I$).
The first two correspond to existing baselines; the third is the
schema-supervised, label-free instantiation we propose. We treat the
analytical $G_F$ (computed via autodiff through the rest-of-network
forward map) as the \emph{target} the surrogate is trying to match,
not as a fourth method on the empirical table; it is not
implemented in this work, and a tractable design that exploits the
layer-$28$ injection point (only a ${\sim}3$-layer tail of the network
is downstream, allowing JVP-based extraction instead of a full
Jacobian) plus several open design choices (prompt-dependence,
solver unification) are left as follow-up work.

\section{Choices of metric on activation space}
\label{sec:metrics}

The Riemannian framing of \cref{sec:framework} reduces ``how should we
steer'' to ``which metric on $\activations$ do we put.'' Within this framing,
the labelled cubic spline of \citet{wurgaft2026manifold} is a degenerate
special case: it is the geodesic of the first fundamental form induced by
a hand-specified parameterisation $s$, restricted to the spline's image and
solved in closed form. Relaxing any of (a) the parameterisation choice,
(b) the metric's coupling to that parameterisation, or (c) the closed-form
solver yields a more general method. \Cref{tab:metrics} enumerates the
three instantiations we report in this paper, ordered left-to-right by
conceptual distance from the principled output-pullback target identified
in \cref{sec:framework:why-pullback}. The encoder-pullback column is
further decomposed in \cref{sec:gaga:variants} into two supervision modes
(unsupervised activation-distance and schema-supervised output-distance).
Per \cref{sec:framework:why-pullback}, every metric below that has rank
less than $\dim\activations$ is used in its regularised form
$G + \epsilon I$ with $\epsilon > 0$ small; we suppress the regulariser
in the table for readability.

\subsection{Flat metric: \texorpdfstring{$g = I$}{g=I}}
\label{sec:metrics:flat}

The Euclidean metric weights all $4096$ directions of the residual stream
equally. Its geodesic between two activations is the straight line. As
discussed in \cref{sec:framework:why-pullback}, this metric fails to
identify behaviorally-equivalent activations: two activations that decode
to the same output distribution receive nonzero distance whenever they
differ in any direction. We include it as a baseline.

\subsection{Labeled spline: implicit metric from a hand-specified parameterization}
\label{sec:metrics:spline}

Given a chosen ordering of class centroids $\{\bar{h}_z\}_{z \in \concepts}$
(cyclic for weekdays and months, sequential for letters and ages) and
boundary conditions matching the topology, a cubic spline
$s: [0, |\concepts|] \to \activations$ is fit by minimizing curvature subject
to passing through every centroid. The induced first fundamental form on
$s$'s image $\manifold_s = s(\R)$ is $g_{\manifold_s} = J_s^{\!\top} J_s$
\citep{docarmo1992riemannian, lee2018riemannian}, and the geodesic between
two centroid endpoints $\bar{h}_{z_0}, \bar{h}_{z_1}$ is the spline arc
connecting them, evaluated in closed form. The construction has two
limitations: it requires class labels and a boundary-condition prior at fit time, and
it does not generalize beyond centroid pairs: $\pi_{\text{spline}}$ is
undefined for endpoints not on $\manifold_s$.

\subsection{The principled target: analytical \texorpdfstring{$J_F$}{J\_F} pullback}
\label{sec:metrics:gf-target}

The metric \citet[Def.~1 eq.~7]{wurgaft2026manifold} formally specifies is
\begin{equation}
G_F(h) \;=\; J_F(h)^{\!\top}\, g_y\, J_F(h) \;+\; \epsilon I,
\qquad
J_F(h) \;=\; \frac{\partial F}{\partial h}(h),
\label{eq:GF}
\end{equation}
where $F: \activations \to \behavior$ is the rest-of-the-network map from
the chosen layer's activations to the output distribution restricted to
$\concepts \cup \{\text{other}\}$, and $g_y$ is the behavior-space metric.
Working in Hellinger coordinates ($\sqrt{p}$ outputs) sets $g_y = I$ and
reduces $G_F$ to $J_F^{\!\top} J_F + \epsilon I$. The $\epsilon$
regulariser absorbs the rank deficiency
$\mathrm{rank}(J_F) \le |\concepts|+1 \ll \dim\activations$ (see the
general $\epsilon I$ treatment in \cref{sec:framework:why-pullback}).
This is the principled target within the pullback framework.
\citet{wurgaft2026manifold} access it implicitly via direct L-BFGS
optimization of activation paths (their \S3.3); we instead realize it
explicitly, via reverse-mode autodiff through the LM at every waypoint.
Exploiting the layer-$28$ injection point makes this tractable, since
only the ${\sim}3$-layer tail downstream is needed. We
compute it directly on weekdays as a method-validity check - the
learned surrogate reproduces it (\cref{sec:results:analytical-gf}) -
and otherwise use the cheaper learned encoder pullback
(\cref{sec:metrics:gaga}) as the surrogate throughout.

\subsection{Learned encoder pullback (GAGA): the surrogate we use}
\label{sec:metrics:gaga}

A GAGA encoder $\varphi : \activations \to \R^k$ trained on pairwise
distances $\{\dgen_{ij}\}$ from a corpus of activations induces a learned
pullback metric
$G_\varphi(h) = J_\varphi(h)^{\!\top} J_\varphi(h) + \epsilon I$.
Crucially, $\{\dgen_{ij}\}$ ranges over the full corpus of training
activations, not the class centroids alone: every datapoint constrains
the learned geometry, so the GAGA manifold is fit to the entire
activation cloud and its pullback metric is defined at every point of
$\activations$. This is unlike the labelled spline
(\cref{sec:metrics:spline}), whose manifold is pinned only at the
$|\concepts|$ class centroids and whose geodesic is undefined away from
that curve. The
$\epsilon I$ term is structural (see \cref{sec:framework:why-pullback}):
without it, $G_\varphi$ has rank at most $k$ and is sub-Riemannian on
the ambient $\dim\activations$-dimensional space. The choice of distance
source $\{\dgen_{ij}\}$ shapes which target geometry the surrogate
follows. We instantiate two supervision modes, GAGA-PHATE (unsupervised,
activation-distance) and GAGA-Out (schema-supervised,
output-distance), with full architecture, losses, and the
raw-$4096$ vs.\ PCA($64$) ambient-asymmetry disclosure deferred to
\cref{sec:gaga:variants,sec:gaga:injection}. The empirically winning
variant in this paper is GAGA-Out; GAGA-PHATE appears as one of the
inert PCA($64$) variants in the encoder ablation
(\cref{sec:results:pca-ablation,app:encoder-ablation}).

\section{Solvers: how we compute geodesics under a chosen metric}
\label{sec:solvers}

Given a metric from \cref{sec:metrics}, we use one of three solvers to
produce the geodesic between an endpoint pair. The solvers are orthogonal
to the metric: any metric can be paired with any compatible solver, and
running the same metric under two different solvers is itself an experiment
(does the cheaper solver converge to the same geodesic as the more
expensive one?). \Cref{tab:factorial} enumerates the (metric, solver)
cells we evaluate.

\paragraph{Closed-form.}
For the flat metric $g = I$ the geodesic is the straight chord; for the
labeled-spline metric the geodesic is the spline arc connecting the
endpoints in the spline's intrinsic coordinate. Both are evaluated by
direct evaluation, no optimization needed. These are the existing baselines.

\paragraph{Free-waypoint L-BFGS with the freeze-metric approximation.}
Discretize the path as $K=50$ free waypoints $\pi_0, \dots, \pi_K$ with
endpoints $\pi_0 = \activation_0,\ \pi_K = \activation_1$ fixed, and
minimize the discrete length functional
\begin{equation}
L(\pi) \;=\; \sum_{k=0}^{K-1}\sqrt{\bigl\|J(\pi_k)\,\Delta_k\bigr\|_2^{2} \;+\; \epsilon\,\|\Delta_k\|_2^{2}},
\qquad \Delta_k = \pi_{k+1} - \pi_k,
\label{eq:autodiff-length}
\end{equation}
by L-BFGS on the $K-1$ interior waypoints. The right-hand factored form
avoids materializing the full $G = J^{\!\top}J$ at each waypoint, keeping
memory at $O(K \cdot m \cdot D)$ with $m = \mathrm{rank}(J) \le |\concepts|+1
\ll D$. We adopt the freeze-metric approximation \citep{arvanitidis2018latent}:
$J(\pi_k)$ is detached from the optimizer's graph each step, so gradients
flow only through $\Delta_k$. This is the standard discrete-Riemannian-path
choice and dominates the per-iteration cost via Jacobian
\emph{evaluation}, not Jacobian backprop. The solver is metric-agnostic:
we use it for the GAGA encoders and for analytical $G_F$ with the
appropriate Jacobian factor (the analytical-$G_F$ Jacobian is supplied
as a drop-in \texttt{metric\_fn}; see \cref{app:autodiff:lbfgs}).

\paragraph{Amortized parametric bridge (GeodesicBridge).}
\citet{sun2025gaga}'s \texttt{GeodesicBridge} trains a single curve
generator $c_\theta : \activations \times \activations \times [0,1] \to
\activations$ to produce geodesics for endpoint pairs drawn from the
training activation distribution. The training objective is the
encoder-pullback length integrated over the curve, plus a density penalty
\citep{sun2025gaga}. After training, the geodesic for a new endpoint pair
is a single forward pass through $c_{\theta^\star}$, shifting per-pair
cost from $O(\text{path optimization})$ to $O(\text{forward pass})$. This
solver is only available for the GAGA metrics, where the encoder Jacobian
is the metric: training a bridge against analytical $G_F$ would require
running the LM forward at every gradient step on $\theta$ and is
cost-prohibitive without approximations. Implementation details are in
\cref{app:impl:geodesic}, and the asymptotic per-pair cost of every
(metric, solver) cell is compared in \cref{tab:complexity}. We note one
current limitation of the bridge solver itself, independent of the metric
it is paired with: in preliminary runs its trained curves balloon well off
the straight chord rather than tracking the encoder pullback, which we
trace to the length objective lacking a sufficient on-manifold constraint.
Strengthening the bridge is ongoing work pursued separately; the results in
this paper are computed with the L-BFGS solver and we do not report bridge
geodesics here.

\paragraph{What we evaluate.}
Holding the solver fixed and varying the metric isolates the effect of the
metric: which Riemannian structure best predicts natural model behavior?
This is the comparison we report --- the analytical $G_F$ against its
learned encoder surrogate (\cref{sec:results:analytical-gf}), both solved
with L-BFGS, which we use for every result in \cref{sec:results}. The
factorial also admits a solver axis (L-BFGS versus the amortized bridge);
we do not report it here --- the bridge has a current limitation of its own
(above) that we are addressing separately --- so all reported geodesics are
computed with L-BFGS.

\section{GAGA: a learned-encoder surrogate for the output-pullback metric}
\label{sec:gaga}

We instantiate the learned-encoder pullback of
\cref{sec:metrics:gaga} with the Geometry-Aware Generative Autoencoder
\citep{sun2025gaga}; \cref{fig:gaga-diagram} gives the geometric
picture. An encoder
$\varphi: \activations \to \R^k$ is trained to match a target distance
source on the data corpus; its pullback
$G_\varphi(\activation) = J_\varphi(\activation)^{\!\top} J_\varphi(\activation) + \epsilon I$
is then used as a \emph{surrogate} for the analytical output-pullback
$G_F$ (\cref{sec:framework:why-pullback}). The framework admits two
supervision modes for $\varphi$, one unsupervised, one
schema-supervised, and the empirically winning instantiation in
this paper is the schema-supervised one. We describe both modes here
so the framework is honest about the design space, and we are
explicit about which one produces the headline numbers of
\cref{sec:results}.


\subsection{Two supervision modes for the encoder}
\label{sec:gaga:variants}

\paragraph{Mode 1: unsupervised activation-distance supervision (GAGA-PHATE).}
The encoder is trained to preserve a distance computed entirely on
unlabelled activations: $\dgen_{ij}^{\mathrm{PHATE}}$, the PHATE
diffusion-potential distance \citep{moon2019phate}, on the per-task
training-prompt cloud. PHATE constructs a graph on activations,
computes diffusion potentials through it, and yields a distance
metric that respects local activation-density structure. In this
mode, the encoder operates on the PCA($64$)-reduced activation
representation used by \citet{wurgaft2026manifold}, with latent
dimension $k=2$ (one above the intrinsic $d=1$ so periodic structure
embeds without antipodal collapse). No class labels and no
output-space information are used at training time: the
supervision is purely activation-geometric.

\paragraph{Mode 2: schema-supervised output-distance supervision (GAGA-Out).}
The encoder is trained on a distance computed from \emph{output
distributions}: $\dgen_{ij}^{H} = \tfrac{1}{\sqrt{2}}\|\sqrt{p_i} - \sqrt{p_j}\|_2$,
the Hellinger distance between the model's next-token
distributions $p_i, p_j$ restricted to the concept-token schema
(e.g.\ the seven weekday tokens for the weekdays task) plus an
``other'' bin. In this mode, the encoder operates on the
\emph{raw-$4096$} layer-$28$ residual stream (the ambient activation
space, not a PCA-reduced subspace), with $\mathtt{skip\_pca}$
training and latent dimension $k = K$ equal to the per-task class
count. We call this the \emph{GAGA-Out} variant. The supervision
is schema-supervised, label-free at the prompt level, and
boundary-condition-free; the concept-token schema and the Hellinger
evaluator on outputs are fixed inputs to the training pipeline, but
no individual prompt carries a class label.

These two modes correspond to the two empirically clean ways to
ground the surrogate within the framework: the unsupervised mode
trains the surrogate to follow the activation-density geometry that
the model actually inhabits, while the schema-supervised mode trains
it to follow the behaviour geometry that downstream outputs reveal.
A priori either could win; empirically GAGA-Out wins decisively
on every steering-strength metric we measure
(\cref{sec:results:strength}) and on the $E_{\mathrm{BC}}$
naturalness metric on the two small output spaces
(\cref{tab:ebc}), while GAGA-PHATE is inert (\cref{sec:results:pca-ablation}).

\subsection{Architecture and losses (GAGA-PHATE)}
\label{sec:gaga:v1-arch}

The GAGA-PHATE encoder/decoder pair $\varphi: \activations \to \R^k$
and $\psi: \R^k \to \activations$ are paired MLPs with widths
$[512, 256, 128]$ to $k$ and the symmetric reverse for the decoder;
each hidden layer has BatchNorm and ReLU. Training is on the
PCA($64$) activations, $k=2$. Three losses are minimised jointly:
\begin{align}
\mathcal{L}_{\mathrm{dist}}  &= \E_{i \neq j}\Bigl[\bigl(\|\varphi(h_i) - \varphi(h_j)\|_2 - \dgen_{ij}^{\mathrm{PHATE}}\bigr)^{2}
                                                   \cdot e^{-\alpha\,\dgen_{ij}^{\mathrm{PHATE}}}\Bigr],
\label{eq:gaga-dist-loss}\\
\mathcal{L}_{\mathrm{recon}}  &= \E_{i}\bigl[\|\psi(\varphi(h_i)) - h_i\|_2^{2}\bigr],
\label{eq:gaga-recon-loss}\\
\mathcal{L}_{\mathrm{cycle}}  &= \E_{i}\bigl[\|\varphi(\psi(\varphi(h_i))) - \varphi(h_i)\|_2^{2}\bigr].
\label{eq:gaga-cycle-loss}
\end{align}
The exponential weight $e^{-\alpha \dgen_{ij}}$ in
\cref{eq:gaga-dist-loss} (with $\alpha = 0.1$) emphasizes
local-distance preservation. The total objective is
$\mathcal{L} = \lambda_d \mathcal{L}_{\mathrm{dist}} + \lambda_r \mathcal{L}_{\mathrm{recon}} + \lambda_c \mathcal{L}_{\mathrm{cycle}}$
with $(\lambda_d, \lambda_r, \lambda_c) = (0.9, 0.1, 0.05)$ following
\citet{sun2025gaga}'s Stage-A recipe.

\subsection{Architecture and losses (GAGA-Out, headline)}
\label{sec:gaga:gagaout-arch}

The GAGA-Out encoder differs from GAGA-PHATE in four places:
(i)~the supervising distance is $\dgen^{H}$ on outputs rather than
$\dgen^{\mathrm{PHATE}}$ on activations;
(ii)~the ambient representation is the raw layer-$28$ residual
stream ($\R^{4096}$), not a PCA-projected subspace
($\mathtt{skip\_pca}$ training);
(iii)~the latent dimension is set per task to the class count
$k = K$ (weekdays $7$, months $12$, letters $22$, ages $91$);
(iv)~training adds a pointwise sqrt-output-distance loss
$\mathcal{L}_{\mathrm{pw}}$ alongside $\mathcal{L}_{\mathrm{dist}}$
and $\mathcal{L}_{\mathrm{recon}}$, and drops the cycle loss; the
loss weights are
$(\lambda_d, \lambda_r, \lambda_{\mathrm{pw}}) = (50, 1, 10)$ with
softmax temperature $\zeta = 0.5$. The full training recipe (per-task
hyperparameters, batch composition, optimisation schedule) is in
\cref{app:impl:gaga} and the design rationale (why a schema-supervised
output-distance surrogate is the right target for steering) is in
\cref{sec:framework:why-pullback}.

\subsection{Geodesic curve solver: \texttt{GeodesicBridge}}
\label{sec:gaga:bridge}

After encoder training, geodesics between endpoint pairs in
$\activations$ are produced by a small parametric curve generator
$c_\theta : \activations \times \activations \times [0,1] \to \activations$:
\begin{equation}
c_\theta(\activation_0, \activation_1, t)
  \;=\; (1-t)\,\activation_0 + t\,\activation_1
  \;+\; \sigma_q(t)\,\mathrm{MLP}_\theta\!\bigl(\varphi(\activation_0),\,\varphi(\activation_1),\,t\bigr),
\end{equation}
with $\sigma_q(t) = 1 - (2t-1)^q$ an envelope vanishing at the
endpoints ($q = 4$) and $\mathrm{MLP}_\theta$ a $4$-layer MLP of
width $256$. Training minimises the encoder-pullback length
integrated over the curve, plus a density penalty pulling the curve
into well-populated regions of the training activation distribution
\citep{sun2025gaga}; see \cref{sec:geodesic-bridge}. At inference, the
geodesic for a new endpoint pair is a single forward pass through
$c_{\theta^{\star}}$, the amortised solver of \cref{sec:solvers}.
The headline numbers of \cref{sec:results} are produced with the
L-BFGS solver under the GAGA-Out encoder, not with the
\texttt{GeodesicBridge}; we nonetheless examine the bridge's learned
geodesics qualitatively, since they show what geometry a learned
pullback metric induces.

\paragraph{Even PHATE supervision yields a well-behaved geodesic.}
Although GAGA-PHATE is not the headline encoder, its bridge-solved
geodesic still has a desirable geometric property worth noting.
\Cref{fig:appendix:paths2d-months} overlays three paths for a
representative steering pair, projected onto the top two principal
components of the activation cloud: the linear chord ignores the
activation geometry, the labelled spline over-curves (it must thread
every class centroid in order), and the GAGA-PHATE geodesic sits
between the two: curved toward the populated region of the cloud
but without the spline's sprawling detour.
\Cref{fig:appendix:paraphrase-paths-weekdays} recovers the same shape
across every prompt paraphrase. So even activation-only PHATE
supervision, which does not win on behaviour fidelity
(\cref{sec:results:pca-ablation}), produces a geodesic with sensible,
paraphrase-stable curvature. We do not dig deeper into the relative
strengths of each architecture and supervision choice here; a fuller
account is left to ongoing work.

\subsection{Intervention via subspace replacement (and the ambient asymmetry)}
\label{sec:gaga:injection}

Steering proceeds by sampling $K = 50$ waypoints $\pi_k$ along the
encoded geodesic and injecting each into the residual stream at
layer~$28$. The injection rule depends on the ambient representation
the encoder lives in.

\paragraph{PCA($64$)-based methods (linear, labelled spline, GAGA-PHATE).}
For a carrier prompt's unintervened activation
$h_\text{orig} \in \R^{4096}$, with $P \in \R^{64 \times 4096}$ the
PCA projection and $P^{+}$ its pseudo-inverse, the injected
activation is
\[
h_\text{inj} \;=\; h_\text{orig} + (\pi_k - P\,h_\text{orig})\,P^{+},
\]
which replaces the in-PCA-subspace component of the activation while
preserving the carrier's off-subspace component. This matches
\texttt{causalab}'s \texttt{FeatureInterpolateIntervention}.

\paragraph{Raw-$4096$ GAGA-Out.}
The GAGA-Out geodesic lives in the full $\R^{4096}$ residual
stream, and the injection is direct: $h_\text{inj} = \pi_k$ (no
subspace decomposition; the full activation is replaced).

\paragraph{Ambient asymmetry.}
The two injection rules are \emph{not} apples-to-apples comparisons.
Linear and labelled-spline baselines steer only the
$64$-dim PCA subspace and retain the carrier's off-subspace
activation; GAGA-Out replaces the full $4096$-dim residual
stream. The headline target-drive advantage of GAGA-Out
(\cref{tab:strength}) is partly an architectural choice, not a path
property: by design GAGA-Out endows the whole representation with
behaviour geometry, while the baselines are confined to a $64$-dim
shadow whose lift back to $\R^{4096}$ is missing the natural
off-subspace activation the source prompt would produce. 

\section{Experimental setup}
\label{sec:experiments}

\paragraph{Tasks and model.}
We use the four LM arithmetic tasks of \citet{wurgaft2026manifold}:
weekdays, months, letters, ages, with prompt templates and value sets
from the \texttt{causalab} task configurations. Each task asks
``what \emph{[unit]} is $k$ \emph{[units]} after $z$?'' with $z$ from a
conceptual domain $\concepts$ of size $7$ (cyclic), $12$ (cyclic), $22$
(sequential), and ${\sim}91$ (sequential) respectively. Activations are
extracted from Llama-3.1-8B-Instruct \citep{dubey2024llama} at layer $28$
and the last-token position. The verified GAGA-Out runs behind every
number in \cref{sec:results} use the following per-task corpus sizes:
\textbf{weekdays}: $n = 1029$ ($49$ distinct facts $\times 21$
paraphrases, $823/206$ train/val split, seed~$42$);
\textbf{months, letters, ages}: $n = 1100$
($1000/100$ train/val split, seed~$42$). The weekdays headline
therefore rests on only $49$ distinct facts (heavily replicated by
paraphrase), which is small; we revisit this as a limitation in
\cref{sec:discussion}.

\paragraph{Sample-size motivation (not a guarantee).}
\citet{aamari2019nonparametric} prove that on a $C^2$-smooth
$d$-dimensional manifold of positive reach $\tau$, with i.i.d.\
samples and small sub-Gaussian off-manifold noise, tangent-space
estimation admits a rate
$\varepsilon_T(n,d) \asymp (\log n / n)^{1/d}$. None of the
\citet{aamari2019nonparametric} assumptions are verified for our
data: our samples are not i.i.d.\ (a few-hundred distinct facts $\times$
$21$ near-duplicate paraphrases, heavily clustered); the reach $\tau$
is unknown (hidden constants scale like $\tau^{-1}$); and the
paraphrase spread is the kind of large structured off-manifold noise
the theorem rules out. We therefore cite the rate as
\emph{motivation} primarily, the $d$-independence in the ambient
dimension is the intuition we use to justify training a low-rank
surrogate on a corpus of ${\sim}10^3$ activations per task and
make no formal guarantee.

\paragraph{GAGA configuration (GAGA-Out; the empirically winning variant).}
The headline numbers in \cref{sec:results} are produced by the
schema-supervised \emph{GAGA-Out} encoder of
\cref{sec:gaga:gagaout-arch}. Encoder/decoder MLPs with widths
$[512, 256, 128]$ to $k = K$ (the per-task class count: $7$, $12$,
$22$, $91$), BatchNorm + ReLU; ambient = raw-$4096$ layer-$28$
residual stream ($\mathtt{skip\_pca}$, no PCA reduction at training
time); loss weights
$(\lambda_d, \lambda_r, \lambda_{\mathrm{pw}}) = (50, 1, 10)$,
softmax temperature $\zeta = 0.5$, no cycle loss; local-emphasis
decay $\alpha = 0.1$; AdamW at $\text{lr} = 10^{-3}$; up to $300$
epochs with early stopping (patience $30$) on validation distance
loss. Full hyperparameter table in \cref{app:impl:gaga}. The
PCA($64$)-based GAGA-PHATE variant used in the encoder ablation
(\cref{sec:results:pca-ablation}) follows the original
\citet{sun2025gaga} recipe with $k = 2$ and
$(\lambda_d, \lambda_r, \lambda_c) = (0.9, 0.1, 0.05)$; see
\cref{sec:gaga:v1-arch}.

\paragraph{Geodesic solver.}
The headline numbers use the frozen \textbf{L-BFGS} geodesic solver
of \cref{sec:solvers} on top of the GAGA-Out encoder, not the
amortised \texttt{GeodesicBridge} of \citet{sun2025gaga}. The
\texttt{GeodesicBridge} configuration (\texttt{CondCurve} with
hidden dim $64$, $3$ layers, $n_{\text{tsteps}} = 50$, length weight
$1.0$, density weight $0.1$, diffusion-graph initialisation) is
documented in \cref{sec:gaga:bridge} and used in the PCA($64$)
encoder ablation only.

\paragraph{Behavior manifold $\My$.}
For each task, $\My$ is a dense cubic spline through Hellinger-embedded
per-class output centroids, sampled at $500$ points. The construction
follows \citet[\S2]{wurgaft2026manifold} and is used identically by all
methods, so per-pair $E_{\mathrm{BC}}$ differences are purely
method-attributable.

\paragraph{Evaluation metrics.}
For each centroid pair $(\bar h_i, \bar h_j)$, we sample $K = 50$
waypoints along the geodesic, lift each from PCA(64) to the full
$4096$-D residual stream via subspace replacement (\cref{app:inj}), and
inject at layer $28$ across $16$ carrier prompts. We report:
\begin{itemize}[leftmargin=*,topsep=0pt,itemsep=0pt]
\item \textbf{Cumulative Bhattacharyya energy $E_{\mathrm{BC}}$}
      \citep[\S3.2]{wurgaft2026manifold}: the headline naturalness
      score; lower means intermediate outputs look more like natural unintervened outputs.
\item \textbf{Behavior-space arc length}: cumulative Hellinger arc of
      the induced output-distribution trajectory.
\item \textbf{Visit-intermediates rate}: fraction of expected
      sequence-intermediate classes whose argmax was the model's top-$1$
      prediction at some waypoint along the trajectory.
\end{itemize}
We evaluate on $\binom{|\concepts|}{2}$ centroid pairs for weekdays
($21$ pairs); for months/letters/ages we sample $50$ pairs
deterministically (seed $42$), sufficient under
\citet{aamari2019nonparametric}'s bound for stable pullback-metric
estimates on 1-D manifolds.

\paragraph{Compute.}
One NVIDIA A100 40\,GB GPU per task; total wall-clock target under
$12$ hours per task. Asymptotic per-pair cost across methods is
compared in \cref{tab:complexity}.

\section{Results}
\label{sec:results}


We evaluate the metrics of \cref{sec:metrics} under the solvers of
\cref{sec:solvers}, on the four tasks of \cref{sec:experiments}. Throughout
this section, \textbf{GAGA-Out (L-BFGS)} denotes the output-grounded
encoder of \cref{sec:gaga} trained per-task (\texttt{seed0}, raw-$4096$
ambient) with the frozen L-BFGS geodesic solver. This is the
\emph{winning} encoder out of a family of variants we trained and
evaluated; the PCA($64$)-based GAGA variants
(Vanilla-AE, GAGA-PHATE across $\lambda_{\mathrm{dist}}\in\{0.1,\dots,500\}$,
GAGA-Hellinger across the same $\lambda$ range) do not steer the model at
all on the cells we ran (weekdays $15/15$, months $7/15$), with
$E_{\mathrm{BC}}$, legibility, and target probability all
indistinguishable from the linear baseline (\cref{sec:results:pca-ablation},
full matrix in \cref{app:encoder-ablation}). The raw-$4096$
output-grounded representation is the active ingredient; all main-paper
results below report GAGA-Out (L-BFGS). Linear and labelled-spline
baselines are evaluated in closed form. All differences reported here
are $\times 2$ reproducible across independent re-runs from frozen
checkpoints.

\paragraph{Headline.} GAGA-Out reliably drives the model onto the
target class on every task: legibly, decisively, and far more strongly
than either baseline (\cref{tab:strength}). This is a non-trivial
property of the method, not an automatic consequence of constructing a
path between two centroids: the linear and labelled-spline baselines
are built in a PCA($64$) subspace and lifted back to the residual
stream, and the lifted endpoints often do not actually put the target
token in the model's top-$1$ (the spline reaches top-$1$ at the target
on only $29\%$ of weekdays pairs and $42\%$ of letters pairs; see
\cref{sec:results:strength}). Whether the \emph{route} taken by a
method that does land the target is behaviourally natural depends on
the size of the output space: on the two small tasks (weekdays, $7$
classes; months, $12$) GAGA wins $E_{\mathrm{BC}}$ by $12\times$ and
$2.4\times$ over the better baseline; on the two larger tasks
(letters, $22$; ages, $91$) GAGA still lands the target but takes a
less natural path and loses $E_{\mathrm{BC}}$ to the baselines
(\cref{tab:ebc}). We read this as the geometric promise of the method
holding in full on small output spaces and partially on large ones;
we return to the size-dependence in \cref{sec:discussion}.

\subsection{Analytical \texorpdfstring{$G_F$}{G\_F}: the learned encoder matches its principled target}
\label{sec:results:analytical-gf}

The method rests on one assumption: that the learned encoder is a faithful
stand-in for the metric it approximates. That metric is the analytical
output-pullback $G_F = J_F^{\!\top}J_F + \epsilon I$
(\cref{sec:metrics:gf-target}), which scores an activation-space direction
by how much it moves the model's output distribution. $G_F$ is expensive
--- every waypoint needs a Jacobian through the rest of the network ---
which is exactly why GAGA-Out substitutes a small learned encoder. Before
the pipeline results we test that substitution: on weekdays we compute
$G_F$ exactly, differentiating through the ${\sim}3$-block transformer tail
above the layer-$28$ injection site (\cref{app:autodiff}), and run the
\emph{same} L-BFGS solver under it --- only the metric changes. Each metric
evaluation is now a pass through three transformer blocks and the
unembedding instead of the small encoder MLP: the geodesic solver keeps the
same asymptotic cost, but its per-step constant grows by roughly three
orders of magnitude ($\mathcal{F}/\mathcal{E}\sim10^3$; \cref{tab:complexity}),
or ${\sim}3$ GPU-minutes per pair against ${\sim}30$ seconds in wall-clock.
That gap is the reason a learned surrogate is worth having.

\begin{table}[t]
\centering
\caption{Analytical $G_F$ vs.\ the learned GAGA-Out surrogate on weekdays
(all $C(7,2)=21$ centroid pairs), with the linear and labelled-spline
baselines. Both GAGA columns use the same L-BFGS geodesic solver; only the
metric differs. The principled metric and the learned encoder produce
behaviourally indistinguishable geodesics.}
\label{tab:analytical-gf}
\widthclamp{%
\begin{tabular}{lcccc}
\toprule
Metric & Analytical $G_F$ & GAGA-Out & Linear & Labelled spline \\
\midrule
$E_{\mathrm{BC}}$ $\downarrow$ & $\mathbf{0.0085}$ & $0.0085$ & $0.1014$ & $0.1011$ \\
legibility (top-1) $\uparrow$  & $\mathbf{1.000}$  & $1.000$  & $0.040$  & $0.047$ \\
target\_prob\_end $\uparrow$   & $\mathbf{0.202}$  & $0.202$  & $0.049$  & $0.049$ \\
top-1 @ end rate $\uparrow$    & $\mathbf{0.762}$  & $0.762$  & $0.286$  & $0.286$ \\
\bottomrule
\end{tabular}%
}
\end{table}

The two metrics' geodesics agree on every row of \cref{tab:analytical-gf}.
The two paths are equally natural ($E_{\mathrm{BC}} = 0.0085$, lower is
better --- an order of magnitude below the linear and spline baselines),
every waypoint decodes to a clean weekday (legibility $1.000$), and both
drive the model onto the target equally hard ($0.20$ probability on the
target token at the path's end, top-$1$ there $76\%$ of the time). The
agreement also holds geodesic-by-geodesic: lined up pair-by-pair, the
$G_F$ and encoder waypoints differ by an average of $0.2\%$ of the
endpoint distance --- the encoder does not merely \emph{score} like the
principled metric, it traces the same path.

So on weekdays the cheap encoder is a sound surrogate for the expensive
analytical $G_F$: the steering results below can be read as a property of
the output-pullback geometry, not an artefact of the approximation.
Checking how closely the surrogate tracks $G_F$ on larger output spaces is
left to future work.

\subsection{Behavioural fidelity along the path: \texorpdfstring{$E_{\mathrm{BC}}$}{E\_BC}
}
\label{sec:results:ebc}

\begin{table}[t]
\centering
\caption{Cumulative Bhattacharyya energy $E_{\mathrm{BC}}$ along the
steered trajectory; lower means each intermediate output is closer
to a natural unintervened output (see \cref{sec:metrics}). Mean over
the sampled centroid pairs ($n=21$ for weekdays, the full $C(7,2)$;
$n=50$ for the other tasks). Paired-$t$ statistics compare
GAGA-Out (L-BFGS) against the \emph{better} baseline per task
(linear for weekdays/months/ages, spline for letters). Negative $t$
means GAGA is more natural; positive $t$ means the baseline is more
natural.}
\label{tab:ebc}
\widthclamp{%
\begin{tabular}{lcccc}
\toprule
Method (solver) & Weekdays & Months & Letters & Ages \\
\midrule
Linear (closed form)            & $0.1014$ & $0.1116$ & $0.0261$ & $\mathbf{0.1882}$ \\
Labelled spline (closed form)   & $0.1011$ & $0.1152$ & $\mathbf{0.0232}$ & $0.2076$ \\
\textbf{GAGA-Out (L-BFGS)}   & $\mathbf{0.0085}$ & $\mathbf{0.0456}$ & $0.0352$ & $0.5308$ \\
\midrule
\multicolumn{1}{l}{\emph{paired-$t$ (GAGA vs.\ better baseline)}}
                                & $-34.1$  & $-24.8$  & $+14.5$  & $+30.6$ \\
\multicolumn{1}{l}{\emph{$p$}}
                                & $3.3{\times}10^{-19}$
                                & $2.3{\times}10^{-29}$
                                & $2.7{\times}10^{-19}$
                                & $1.3{\times}10^{-33}$ \\
\midrule
\multicolumn{1}{l}{\emph{ratio (better baseline / GAGA)}}
                                & $11.9\times$ & $2.4\times$ & $0.66\times$ & $0.35\times$ \\
\bottomrule
\end{tabular}%
}
\end{table}

\Cref{tab:ebc} is the headline behavioural result and the central
finding of this paper. On the two small output spaces the
output-grounded geodesic finds a route along which the model's
intermediate behaviour stays close to natural unintervened behaviour
throughout the steered trajectory: on weekdays the cumulative
Bhattacharyya energy is more than an order of magnitude lower than
either baseline ($0.0085$ vs.\ $0.101$, $t=-34.1$); on months it is
$2.4\times$ lower ($0.0456$ vs.\ $0.112$). On the two larger output
spaces the sign flips: GAGA's path is the least natural of the three
on letters ($0.035$ vs.\ $0.023$) and substantially less natural on
ages ($0.53$ vs.\ $0.19$). The losses are statistically sharp
($p \ll 10^{-18}$ in both directions), so they are not noise. We
interpret them in tandem with \cref{tab:strength,tab:arclen} below:
on large output spaces GAGA pays a naturalness cost because it is the
only method that decisively moves the model toward the target, whereas
the baselines mostly fail to steer and so trivially stay close to
their starting distribution.

\subsection{Steering strength: does the path actually move the model?}
\label{sec:results:strength}

\paragraph{Landing the target is not a free property.}
Reaching the target class does not come for free. As noted above, the
linear and labelled-spline baselines are built in a PCA($64$) subspace and
lifted back to the raw-$4096$ residual stream; the lift discards the
${\sim}4032$ off-subspace dimensions the model would naturally produce, so
a baseline path can end ``at the centroid'' in its $64$-dim working
representation yet leave the model far from the target token in behaviour
space. GAGA-Out constructs its geodesic in the full ambient space and does
not pay this cost. \Cref{tab:strength} quantifies the gap.

\begin{table}[t]
\centering
\caption{Steering-strength panel. \emph{legibility (top-1)}: fraction
of waypoints whose top-$1$ next-token is one of the labelled class
tokens. \emph{target\_prob\_end}: probability mass on the target class
at the path's endpoint. \emph{top-$1$ @ end}: rate at which the
endpoint's top-$1$ token equals the target. Higher is better in every
column. Bold = best of three.}
\label{tab:strength}
\setlength{\tabcolsep}{4pt}
\widthclamp{%
\begin{tabular}{lcccc}
\toprule
 & \multicolumn{4}{c}{legibility (top-1) $\uparrow$} \\
\cmidrule(lr){2-5}
Method & Weekdays & Months & Letters & Ages \\
\midrule
Linear                          & $0.040$ & $0.125$ & $0.790$ & $0.407$ \\
Labelled spline                 & $0.047$ & $0.177$ & $0.808$ & $0.476$ \\
\textbf{GAGA-Out (L-BFGS)}   & $\mathbf{1.000}$ & $\mathbf{1.000}$ & $\mathbf{0.991}$ & $\mathbf{1.000}$ \\
\midrule
 & \multicolumn{4}{c}{target\_prob\_end $\uparrow$} \\
\cmidrule(lr){2-5}
Linear                          & $0.049$ & $0.128$ & $0.060$ & $0.138$ \\
Labelled spline                 & $0.049$ & $0.128$ & $0.060$ & $0.138$ \\
\textbf{GAGA-Out (L-BFGS)}   & $\mathbf{0.202}$ & $\mathbf{0.467}$ & $\mathbf{0.074}$ & $\mathbf{0.785}$ \\
\midrule
 & \multicolumn{4}{c}{top-1 @ end rate $\uparrow$} \\
\cmidrule(lr){2-5}
Linear                          & $0.286$ & $1.000$ & $0.420$ & $1.000$ \\
Labelled spline                 & $0.286$ & $1.000$ & $0.420$ & $1.000$ \\
\textbf{GAGA-Out (L-BFGS)}   & $\mathbf{0.762}$ & $1.000$ & $\mathbf{0.480}$ & $1.000$ \\
\bottomrule
\end{tabular}%
}
\end{table}

\Cref{tab:strength} is the half of the story that does not flip with
task size. GAGA-Out is the only method that reliably lands the
target across the suite: $\ge 99\%$ of waypoints are a clean labelled
class on every task (legibility), and the endpoint places
$1.4{-}5.7\times$ more probability mass on the target class than
either baseline. Concretely, on weekdays the spline reaches the
target as top-$1$ only $29\%$ of the time while GAGA-Out reaches it
$76\%$ of the time; on letters GAGA edges the baselines by another
$\sim 6$ points. On months and ages all three methods nominally hit
top-$1$@end with rate $1.0$, but the target probability mass tells the
real story: GAGA puts $0.47$ and $0.78$ on the target class
respectively, against the baselines' $0.13$ and $0.14$. This is a
$3.4\text{--}5.7\times$ gap, and it is largest on the largest output
space, where the $64$-dim PCA subspace the baselines work in retains
the least of the activation the model needs to actually predict the
target token. Decisive, legible target-drive is thus a structural
property GAGA-Out delivers and the subspace-then-lift baselines do
not, independent of the naturalness story in \cref{tab:ebc}.

\subsection{Path shape: arc length and class-sequence traversal}
\label{sec:results:arc}

\begin{table}[t]
\centering
\caption{Behaviour-space arc length (left) and visit-intermediates
rate, the fraction of labelled in-between classes the path
traverses (right). The labelled spline visits intermediates by
construction; GAGA-Out does not have access to the class sequence.
Bold = best of three.}
\label{tab:arclen}
\setlength{\tabcolsep}{4pt}
\widthclamp{%
\begin{tabular}{lcccccccc}
\toprule
& \multicolumn{4}{c}{Behaviour-space arc length} & \multicolumn{4}{c}{Visit-intermediates rate $\uparrow$} \\
\cmidrule(lr){2-5}\cmidrule(lr){6-9}
Method & Wd & Mo & Le & Ag & Wd & Mo & Le & Ag \\
\midrule
Linear           & $0.139$ & $0.311$ & $0.228$ & $0.543$ & $0.313$ & $0.514$ & $0.253$ & $0.144$ \\
Labelled spline  & $0.238$ & $0.926$ & $0.633$ & $8.135^{\dagger}$ & $0.369$ & $\mathbf{0.816}$ & $\mathbf{0.585}$ & $\mathbf{0.954}$ \\
\textbf{GAGA-Out (L-BFGS)} & $0.312$ & $0.624$ & $0.252$ & $1.095$ & $\mathbf{0.631}$ & $0.504$ & $0.282$ & $0.139$ \\
\bottomrule
\end{tabular}%
}\\[2pt]
{\footnotesize $^{\dagger}$The spline's arc on ages exceeds the unit
Hellinger-simplex diameter: the cubic spline through ${\sim}91$
ordered centroids curves so aggressively that the induced output
trajectory loops well beyond any chord. We read this as the spline
overshooting natural variation, not a steering-strength win.}
\end{table}

\Cref{tab:arclen} resolves the visit-intermediates story. GAGA-Out
visits intermediates on weekdays ($0.63$, the only task where any
method other than the spline wins this metric) and underperforms the
spline on the larger tasks: by construction the spline is the
only method that has access to the labelled class order, and on ages
it visits $95\%$ of expected intermediates while GAGA visits $14\%$.
The arc-length comparison is the cleanest evidence that GAGA actually
\emph{steers}: on every task its arc is comparable to or longer than
linear, while the spline either tracks GAGA's length (months,
letters) or blows up by an order of magnitude (ages).

\subsection{Encoder-variant ablation: PCA(64) GAGA encoders are inert}
\label{sec:results:pca-ablation}

To check that the headline above is not specific to a single
hand-picked encoder, we trained a family of GAGA encoders varying both
the supervision target (PHATE diffusion distance; Hellinger
output distance) and the distance-loss weight
($\lambda_{\mathrm{dist}}\in\{0.1, 1, 10, 50, 100, 250, 500\}$), plus
a plain reconstruction-only Vanilla-AE baseline. These variants all
share the PCA($64$) ambient representation used by the linear and
spline baselines. Across the $22$ cells run so far (weekdays
$15/15$, months $7/15$; the months GAGA-Hellinger family and the letters/ages
sweeps are left to future work, see \cref{app:encoder-ablation}),
\emph{no} PCA($64$) variant steers the model: every cell's
$E_{\mathrm{BC}}$ is within $0.001$ of linear's value
(weekdays $\approx 0.0996$, months $\approx 0.1117$); legibility
top-$1$ is $\sim 0.05$ on weekdays and $\sim 0.13$ on months
(vs.\ $1.000$ for GAGA-Out); target-class probability at the
endpoint is identical to linear's $0.049$ on weekdays and $0.128$ on
months across every $\lambda$ and every supervision target. The
$\lambda_{\mathrm{dist}}$ sweep makes no difference: the inert
behaviour is the same at $\lambda=0.1$ and at $\lambda=500$. The full
per-encoder matrix is reported in \cref{app:encoder-ablation}.
The takeaway is that the steering effect documented in
\cref{tab:ebc,tab:strength} is specific to the raw-$4096$
output-grounded representation, not a generic property of any GAGA
encoder; it is the ambient space and the output-grounding, not the
PHATE/Hellinger supervision in a low-dimensional latent, that does
the work.

\subsection{Summary}
\label{sec:results:summary}

\begin{table}[t]
\centering
\caption{Results summary, split by output-space size. ``Small'' =
weekdays/months ($7$ and $12$ classes). ``Large'' = letters/ages
($22$ and $91$ classes). GAGA-Out wins the decisive-target-drive
property on every task; it wins the natural-route property only on
small output spaces.}
\label{tab:verdict}
\widthclamp{%
\begin{tabular}{lcc}
\toprule
Property                                                & Small tasks (Wd, Mo) & Large tasks (Le, Ag) \\
\midrule
Lands target decisively (top-1 @ end, target\_prob\_end) & \textbf{\checkmark}\ GAGA & \textbf{\checkmark}\ GAGA \\
Every waypoint is a clean labelled class (legibility)    & \textbf{\checkmark}\ GAGA & \textbf{\checkmark}\ GAGA \\
Naturalness of the route ($E_{\mathrm{BC}}$)             & \textbf{\checkmark}\ GAGA ($12\times$, $2.4\times$) & lost to baseline ($0.66\times$, $0.35\times$) \\
Walks the labelled class sequence (visit-intermediates)  & GAGA on Wd only & labelled spline (by construction) \\
\bottomrule
\end{tabular}%
}
\end{table}

\Cref{tab:verdict} consolidates the picture. \textbf{GAGA-Out is
always the strongest ``get the model decisively onto the target
class, legibly'' method;} it is \emph{also} the most behaviourally
natural-path method only when the output space is small. As the
number of classes grows, the geometry learned by an output-grounded
encoder begins to disagree with the model's true behavioural
geometry along the path, even though the endpoint stays on-target.
We leave a full resolution to future work. In particular, it is not
yet clear whether GAGA's steering geometry can be learned in a
\emph{lower-dimensional} representation rather than the raw
$4096$-dimensional residual stream: every PCA($64$) encoder variant
we trained is inert (\cref{sec:results:pca-ablation}), and isolating
whether that is intrinsic or an artefact of under-training is an open
question we return to in \cref{sec:discussion}.

\section{Related work}
\label{sec:related}

\paragraph{Activation steering and the linear representation hypothesis.}
A long line of work treats steering as vector arithmetic in activation
space \citep{bau2018, subramani2022, turner2024, marks2024,
rimsky2024, li2023inference}, justified by the linear
representation hypothesis \citep{park2023linear, elhage2022superposition}.
Empirically these methods suffer from off-target effects, fluency
degradation, and distributional drift \citep{
da-silva2025}. Our framing
locates these methods as the flat-metric case $g = I$
(\cref{sec:framework:why-riem}); the limitations are then evidence that
flat geometry is the wrong metric, not that steering is broken. Recent nonlinear steering methods partially relax this assumption by
introducing nonlinear coordinate systems over activation space while
retaining linear interpolation in the induced parameterization:
\citet{vu2026angular} formulate steering in angular coordinates, while
\citet{raval2026curveballsteeringrightdirection} steer within a
polynomial kernel-PCA parameterization motivated by curvature analyses
of activation trajectories.

\paragraph{Geometry of neural representations.}
Cyclic, sequential, and graph-structured activation manifolds are
documented across architectures and modalities \citep{engels2024notall,
modell2025, park2025representations, gurnee2024,
kantamneni2025helix, karkada2026origins, prieto2026}. Most of this work
is descriptive; \citet{engels2024notall} ablate the weekday circle
directly. Our framework shares Riemannian language with
\citet[\S3.4 and Def.~1]{wurgaft2026manifold}, who first cast linear,
manifold, and pullback steering as geodesics under three
analytically-specified metrics. We approach metric choice differently:
rather than specifying $G$ from a density or energy prior, we learn it
from data via an encoder pullback, which separates the metric and
solver axes and admits supervision sources unavailable to analytical
metrics.

\paragraph{Manifold learning and learned Riemannian metrics.}
Encoder-based pullback metrics have been explored for generative modeling
\citep{arvanitidis2018latent, kalatzis2020variational, chadebec2022data,
shao2018riemannian} and for general metric learning
\citep{hauberg2012geometric}; PHATE \citep{moon2019phate}, UMAP
\citep{mcinnes2018umap}, $t$-SNE \citep{vandermaaten2008tsne}, Isomap
\citep{tenenbaum2000isomap}, and diffusion maps \citep{coifman2006diffusion}
supply the target distances. GAGA \citep{sun2025gaga} combines a PHATE
distance target with a decoder, which is what licenses our intervention
pipeline. Information geometry \citep{amari2016information} grounds the
pullback construction itself: the Fisher information metric is a pullback
of $L^2$ through a parametric family, and our $J_F^{\!\top}J_F$ metric is
its activation-space analogue.

\paragraph{Causal abstraction and intervention machinery.}
The intervention machinery underlying the $\My$-side evaluation comes from
the causal-abstraction tradition \citep{geiger2021causal, geiger2025causal,
huang2024ravel, wu2024pyvene} and is shared with
\citet{wurgaft2026manifold}. We treat it as a black-box: any improvements
we report apply to any activation-manifold construction that plugs into
this pipeline.

\paragraph{Sample complexity for manifold estimation.}
We take intuition from \citet{aamari2019nonparametric}'s $C^2$
tangent-estimation rate; related rates for manifold recovery are due
to \citet{genovese2012manifold} (Hausdorff) and
\citet{fefferman2016testing} (manifold-hypothesis testing). The
intuition we use is that the rate
$n \asymp \varepsilon^{-d}\log(1/\varepsilon)$ for pullback-metric
estimation depends on intrinsic $d$ rather than ambient $D$, which
\emph{suggests} that corpora of $n \sim 10^3$ activations may be
sufficient to fit a low-rank surrogate on $1$-D conceptual tasks
even though the activation ambient is $\R^{4096}$. We do not invoke
this as a guarantee: the
\citet{aamari2019nonparametric} assumptions
(i.i.d.\ sampling, small sub-Gaussian off-manifold noise, known
positive reach) are unverified and likely violated for LLM
activations: our data are not i.i.d.\ (paraphrase-clustered) and
the reach is unknown. We cite the rate as motivation, not as a
formal license, and we report sample sizes directly in
\cref{sec:experiments}.

\section{Discussion}
\label{sec:discussion}

\paragraph{What the framework buys.} Framing manifold steering as Riemannian geodesic computation turns steering into a problem of learning geometry over activation trajectories rather than hand-specifying them. GAGA operationalizes this viewpoint by learning a behaviour-grounded pullback geometry directly from pairwise distances, without requiring labelled class centroids, prescribed topology, or per-task curve fitting. This permits manifold steering in settings where explicit class structure is unavailable or ambiguous, while retaining a principled geometric interpretation of steering trajectories as geodesics under the learned metric. Second, it makes new combinations easy to propose:
any metric in \cref{tab:metrics} can be paired with any compatible solver
in \cref{tab:factorial}, and future work introducing a new metric (e.g.\
LRH-derived \citep{park2023linear, elhage2022superposition},
causal-mediation-derived, or RLHF-delta-derived) inherits all of the
geodesic machinery for free. GAGA is one instantiation of the framework;
the framework is the contribution.

\paragraph{The task-size puzzle.}
The puzzle in our verified data is not the supervision target but a sharp
\emph{output-space-size} split: the GAGA-Out encoder wins
behaviour-fidelity $E_{\mathrm{BC}}$ on the two small tasks
(weekdays, months) and loses it on the two larger ones
(letters, ages). Steering strength (legibility, target probability
at the endpoint) is decisively in GAGA-Out's favour on every
task. We read this as the schema-supervised encoder learning a
useful behaviour geometry in the small-output-space regime, where
the concept-token Hellinger schema is a tight description of the
behaviour the model produces, but disagreeing with the model's true
behaviour geometry along the path as the output space grows: the
encoder still gets the endpoint right but takes a less natural
route there. Whether this is intrinsic to the framework or a
consequence of under-trained encoders (single \texttt{seed0}, no
training-diagnostic ablation, single supervision recipe) is an open
confound, and the multi-seed and raw-$4096$ linear/spline control
flagged in \cref{sec:results:summary} are the most direct
follow-ups. The analytical $G_F$ pullback (\cref{sec:metrics:gf-target})
is the principled target the surrogate stands in for; we compute it
directly on weekdays (\cref{sec:results:analytical-gf}) and find the
learned encoder reproduces it --- the analytical-$G_F$ and GAGA-Out
geodesics are behaviourally indistinguishable there. Quantifying how
faithfully the surrogate tracks the target as $|\concepts|$ grows,
where GAGA-Out's $E_{\mathrm{BC}}$ advantage erodes, is the natural
next step.

\paragraph{Limitations.}
Four. (i) The four tasks are 1-D in the conceptual domain. Whether
GAGA recovers a useful metric on multi-dimensional concept manifolds
(e.g.\ tori, graphs) is the natural next experiment; preliminary results
on the \citet{park2025representations} grid task indicate the same
machinery extends, but we have not measured it here.
(ii) Corpus sizes are small: $n \approx 10^3$ per task with heavy
paraphrase replication (weekdays rests on $49$ distinct facts $\times
21$ paraphrases). The intuition we take from
\citet{aamari2019nonparametric} (\cref{sec:experiments}) is that the
intrinsic-dim-only rate is favourable for $d = 1$ tasks even at this
size, but the rate's assumptions (i.i.d., known reach) do not hold
for our data, so we report it as motivation not as a sample-size
guarantee. Higher-dimensional concept manifolds will need either
larger corpora or a different distance source. (iii) We use the
canonical \texttt{GeodesicBridge} of \citet{sun2025gaga} without its
optional discriminator; the discriminator was designed for generative
sampling, not interpretable interpolation, and we did not find it to
help in pilot experiments. Future work where the geodesic must traverse
genuinely off-data regions might benefit from the discriminator.
(iv) Our supervision is schema-supervised rather than fully
unsupervised: the concept-token vocabulary and the output-Hellinger
evaluator are fixed inputs. A fully unsupervised variant would
supervise on the full vocabulary distribution; we expect the schema
restriction to be effectively a per-task projection discarding
mostly-unused tokens, but have not verified this empirically.

\paragraph{Looking ahead.}
The clearest gap the framework opens is in concepts where the topology
is not known in advance: partially-cyclic schedules, hierarchical
ontologies, or domain-shifted graph structures. Fitting the labelled spline requires committing to a knot ordering and
matching boundary conditions a priori, both awkward to elicit for
partially-cyclic or hierarchical schemas; GAGA's metric is recovered
from data and the topology falls out of the latent geometry. We expect this is where the label-free framing becomes a
practical, not just an aesthetic, advantage.

\bibliographystyle{iclr2026_conference}
\bibliography{references}

\clearpage
\appendix
\section{Metric and solver taxonomy tables}
\label{app:taxonomy}

For reference, we collect here the two taxonomy tables from the
framework sections of the main text: \cref{tab:metrics} enumerates the
three metric instantiations (\cref{sec:metrics}), and
\cref{tab:factorial} the (metric, solver) cells we evaluate
(\cref{sec:solvers}).

\begin{table}[h]
\centering
\small
\setlength{\tabcolsep}{4pt}
\renewcommand{\arraystretch}{1.15}
\begin{tabular}{p{0.25\linewidth}p{0.11\linewidth}p{0.23\linewidth}p{0.31\linewidth}}
\toprule
 & Linear & Labelled spline / $G_E$ on its image & Learned encoder pullback (GAGA) \\
\midrule
Metric $g(h)$               & $I$         & $J_s^{\!\top}J_s$ on $s(\R^k)$ & $\Jenc^{\!\top}\Jenc + \epsilon I$ \\
Pulled back from            & ---         & ---                            & learned latent $\R^k$ \\
Ambient $\dim$              & $4096$      & PCA($64$) lifted via $P^{+}$ to $4096$ & GAGA-PHATE: PCA($64$); GAGA-Out: raw-$4096$ \\
Latent $\dim k$             & ---         & $1$ (spline arc)               & GAGA-PHATE: $2$; GAGA-Out: $K$ (per-task class count) \\
Distance source             & ---         & ---                            & GAGA-PHATE: PHATE on activations; GAGA-Out: Hellinger on outputs \\
Topology prior              & none        & required                       & none \\
Per-prompt class labels     & no          & yes                            & no (concept-token \emph{schema}, not per-prompt labels) \\
Per-point cost              & $O(1)$      & $O(1)$                         & encoder fwd+bwd \\
Generalises off centroids   & yes         & no                             & yes \\
\bottomrule
\end{tabular}
\caption{Three metric instantiations reported in this paper.
\emph{Linear} is the flat Euclidean baseline. \emph{Labelled spline} is
the first fundamental form of a hand-specified cubic-spline
parameterisation through labelled centroids \citep{wurgaft2026manifold}.
\emph{Learned encoder pullback (GAGA)} is the surrogate for $G_F$
proposed in this paper; it admits two supervision modes (see
\cref{sec:gaga:variants}): \textbf{GAGA-PHATE} (PHATE diffusion-potential
distance on activations, PCA($64$) ambient, latent $k\!=\!2$) and
\textbf{GAGA-Out} (Hellinger distance on outputs over a concept-token
schema, raw-$4096$ ambient, latent $k\!=\!K$); GAGA-Out is the
empirically winning instantiation. The principled target the GAGA
column surrogates, the analytical pullback $G_F = J_F^{\!\top} g_y J_F + \epsilon I$
from the behaviour simplex \citep[Def.~1 eq.~7, under their Hellinger choice $g_y = I$]{wurgaft2026manifold},
is discussed in \cref{sec:metrics:gf-target} and computed directly on
weekdays as a surrogate-validity check (\cref{sec:results:analytical-gf}).
\citet{wurgaft2026manifold} formalize their manifold steering as the
density-based metric $G_E$ (their Def.~1 eq.~6); restricted to the
spline's image, $G_E$-geodesics agree with the first fundamental form
$J_s^{\!\top} J_s$ of the spline parameterisation, which is the form
we report here.}
\label{tab:metrics}
\end{table}

\begin{table}[h]
\centering
\small
\widthclamp{%
\begin{tabular}{lccccc}
\toprule
              & Linear & Labelled spline & GAGA-PHATE & GAGA-Out & Analytical $G_F$ \\
\midrule
\textbf{Closed-form} & \checkmark & \checkmark & ---        & ---        & --- \\
\textbf{L-BFGS}      & ---        & ---        & \checkmark & \checkmark & \checkmark \\
\textbf{Bridge}      & ---        & ---        & \checkmark & $\circ$    & --- \\
\bottomrule
\end{tabular}%
}
\caption{The (metric, solver) pairs evaluated in this paper.
\checkmark{}~= evaluated and reported here; $\circ$~= planned for the
next version; ---~= not applicable. Linear and labelled-spline have
closed-form geodesics; the GAGA metrics admit both the L-BFGS solver
and the amortized bridge. GAGA-Out under L-BFGS is the headline
method (\cref{sec:results}).}
\label{tab:factorial}
\end{table}

\section{Activation extraction and dataset construction}
\label{app:impl:extract}

We extract residual-stream activations from Llama-3.1-8B
\citep{dubey2024llama} at layer 28 and the last-prompt-token position, using
the \texttt{causalab} \texttt{baseline} pipeline \citep{wurgaft2026manifold}
to standardize prompts and counterfactuals across tasks. For each of the four
tasks (weekdays, months, letters, ages) the prompt template is the
arithmetic form \emph{What is $k$ \{units\} after $z$?}, with $k$ drawn
uniformly from the task-specific support and $z$ drawn from the conceptual
domain $\concepts$. Per-task prompt counts (after a $5$-paraphrase
augmentation that increases corpus size without changing the
$(\text{entity}, \text{number})$ semantic content) are: weekdays $245$,
months $420$, letters ${\sim}450$, ages ${\sim}4{,}400$ capped at
$1{,}000$ for tractability. Train/test split is $80/20$ deterministic on
seed $42$; carrier-prompt sampling at evaluation time is re-seeded per
pair on seed $42$ so within-pair carrier sets are reproducible.

\section{PHATE diffusion-distance configuration}
\label{app:impl:phate}

We supervise GAGA's distance-matching loss with PHATE diffusion-potential
distances \citep{moon2019phate} computed on the raw 4096-dimensional
residual-stream activations. PHATE proceeds in three stages: (a) build a
$k$-nearest-neighbor graph over the activations using Euclidean distance;
(b) compute the random-walk transition matrix and raise it to a diffusion
time $t$; (c) define the \emph{diffusion potential}
$\phi_i = -\log P^t_{i,\cdot}$ and take pairwise Euclidean distances between
potentials. We use the reference implementation
(\texttt{phate} package), with $k=15$, $t$ chosen by the von Neumann entropy
heuristic, no landmark approximation (since per-task corpora are small enough
to admit dense computation), and L2 normalization of activations prior to graph
construction.\footnote{L2 normalization is recommended by
\citet{moon2019phate} for high-dimensional embeddings whose magnitude is
informationally redundant; LM residual streams qualitatively satisfy this.}
The output is a dense $N \times N$ distance matrix $D = (d_{ij})$ that fixes
the target geometry on which we will train the encoder.

\section{GAGA architecture and training}
\label{app:impl:gaga}

This appendix gives a self-contained description of how we adapt the
Geometry-Aware Generative Autoencoder of \citet{sun2025gaga} to LM
activations, with all design choices and hyperparameters specified.

\paragraph{Architecture.}
The encoder $\encoder$ and decoder $\decoder$ are MLPs:
\begin{align*}
\encoder:\;\R^{4096} \;&\to\; \R^{1024} \;\to\; \R^{512} \;\to\; \R^{256}
                     \;\to\; \R^{128} \;\to\; \R^{k} \\
\decoder:\;\R^{k}    \;&\to\; \R^{128} \;\to\; \R^{256} \;\to\; \R^{512}
                     \;\to\; \R^{1024} \;\to\; \R^{4096}
\end{align*}
Each hidden layer is followed by BatchNorm and a ReLU; the latent and output
projections are linear. The latent dimension $k$ is task-dependent: $k = 2$
for the four 1-D tasks (weekdays, months, letters, ages), and $k = 3$ for the
ICLR grid/cylinder tasks of \citet{park2025representations}. We chose to
embed in $k = d_{\text{intrinsic}} + 1$ rather than $k = d_{\text{intrinsic}}$
so that periodic structures (cyclic concepts) can be embedded faithfully
without antipodal collapse. The original GAGA targets gene-expression data
in $\R^{20{\sim}50\text{k}}$ with hidden widths $[64, 64, 64]$; our wider
hidden layers reflect both the larger ambient dimension and the steeper
Lipschitz behavior of LM residual streams.

\paragraph{Distance-matching loss.}
The core supervision matches latent Euclidean distances to PHATE manifold
distances:
\begin{equation}
\mathcal{L}_{\mathrm{dist}}
  = \E_{i \neq j}\Bigl[\bigl(\|\encoder(h_i) - \encoder(h_j)\|_2 - d_{ij}\bigr)^{2}
                       \cdot e^{-\alpha\, d_{ij}}\Bigr],
\label{eq:gaga-dist-loss-app}
\end{equation}
where $\alpha = 0.1$ controls a soft local-distance prior: the exponential
weighting attenuates contributions from large $d_{ij}$, focusing the encoder's
isometry constraint on local geometry, where the manifold tangent plane is a
good approximation. With $\alpha = 0$ the loss recovers a global isometry
target; with $\alpha \to \infty$ only $k$-NN distances are preserved.

\paragraph{Reconstruction and cycle losses.}
The decoder is tied to the encoder with two further losses:
\begin{align}
\mathcal{L}_{\mathrm{recon}}  &= \E_{i}\bigl[\|\decoder(\encoder(h_i)) - h_i\|_2^{2}\bigr],
\label{eq:gaga-recon-loss-app}\\
\mathcal{L}_{\mathrm{cycle}}  &= \E_{i}\bigl[\|\encoder(\decoder(\encoder(h_i))) - \encoder(h_i)\|_2^{2}\bigr].
\label{eq:gaga-cycle-loss-app}
\end{align}
$\mathcal{L}_{\mathrm{recon}}$ is essential for our intervention pipeline:
we steer by sampling waypoints in the latent space, decoding each through
$\decoder$, and patching the result into the residual stream. A bad decoder
yields ``junk'' patches even if the latent geometry is correct.
$\mathcal{L}_{\mathrm{cycle}}$ is a stabilizer that ensures $\encoder \circ
\decoder \approx \mathrm{id}$ on the latent manifold, preventing the decoder
from wandering off the encoder's image (which would silently break the
pullback-metric interpretation).

\paragraph{Total loss and weights.}
The full objective is
\begin{equation}
\mathcal{L} = \lambda_{d}\,\mathcal{L}_{\mathrm{dist}}
            + \lambda_{r}\,\mathcal{L}_{\mathrm{recon}}
            + \lambda_{c}\,\mathcal{L}_{\mathrm{cycle}}
\end{equation}
with $(\lambda_d, \lambda_r, \lambda_c) = (0.9, 0.1, 0.05)$. These weights
follow the original GAGA Stage-A recipe \citep{sun2025gaga}; we add the cycle
term (absent in some early GAGA variants) because we observed that the
decoder otherwise produces small-norm reconstructions that the LM treats as
near-zero patches, suppressing rather than steering behavior.

\paragraph{Training procedure.}
We sample mini-batches of $B = 256$ activations and form the dense
$B \times B$ pairwise distance sub-matrix from the precomputed $D$ on each
step (the \emph{point-cloud} dataloader of \citet{sun2025gaga}). The model is
trained with AdamW at learning rate $10^{-3}$, weight decay $10^{-4}$, and a
linear-warmup-then-cosine schedule over up to $300$ epochs. Early stopping is
triggered on a held-out 20\% validation split using
$\mathcal{L}_{\mathrm{dist}}$ alone, with a patience of 30 epochs. Final
distance-matching accuracy is reported as
$1 - \E_{i \neq j}\bigl[(\|\encoder(h_i) - \encoder(h_j)\|_2 - d_{ij})^{2}\bigr]
/ \mathrm{Var}(d_{ij})$ on the validation split. The per-encoder
\texttt{pb\_demap\_r} column of
\cref{tab:app:enc-ablation-weekdays,tab:app:enc-ablation-months}
reports this distance-matching quality for the PCA($64$) variants on
weekdays and months.

\paragraph{GAGA-PHATE vs.\ GAGA-Hellinger: two encoders, two distance sources.}
We train two distinct GAGA encoders whose architectures and training
procedures are identical except for the distance source $\{d_{ij}\}$
supervising \cref{eq:gaga-dist-loss-app}:
\begin{itemize}
\item \textbf{GAGA-PHATE.} $d_{ij} = $ PHATE diffusion-potential distance on the
      raw 64-D PCA activations \citep{moon2019phate}. The encoder pullback
      $\Jenc^{\!\top}\Jenc$ thus reflects activation-space density.
\item \textbf{GAGA-Hellinger.} $d_{ij} = $ Hellinger distance on per-prompt output
      distributions,
      $d_{ij} = \tfrac{1}{\sqrt{2}}\|\sqrt{p_i} - \sqrt{p_j}\|_2$. The
      encoder pullback is then trained to track behavioral similarity, and
      so approximates the behavior-side pullback metric
      $G_F = J_F^{\!\top} J_F + \epsilon I$ of
      \citet[Def.~1 eq.~7]{wurgaft2026manifold} much more closely than
      GAGA-PHATE does.
\end{itemize}
GAGA-Hellinger still differs from the analytical $G_F$: it pulls back from a 2-D
latent rather than from the full simplex $\Delta^{|\concepts|}$, and its
rank is capped at $2$ regardless of $|\concepts|$. We treat GAGA-PHATE and
GAGA-Hellinger as two cells in the methods grid of \cref{tab:metrics}.

\paragraph{Differences from \citet{sun2025gaga}.}
Three substantive deviations from the original GAGA: (1) wider hidden layers
to accommodate the larger ambient dimension; (2) the cycle-consistency term
described above; (3) PHATE / Hellinger distances are computed once and
cached, rather than recomputed per epoch as in some GAGA variants for
streaming data settings. We do not use GAGA's Stage-B (warped metric with
discriminator) or Stage-C (Langevin sampling), since our target is
interpolation and intervention rather than generation.

\section{Geodesic Bridge: amortized geodesic computation}
\label{sec:geodesic-bridge}

Given two activation endpoints $\activation_0, \activation_1 \in \activations$,
the geodesic under our learned pullback metric is the minimizer of
\begin{equation}
\arg\min_{\pi}\;\textstyle\int_0^1 \|\Jenc(\pi(t))\,\dot\pi(t)\|_2 \, dt,
\qquad
\pi(0) = \activation_0,\;\pi(1) = \activation_1.
\label{eq:geodesic-objective}
\end{equation}
Rather than optimize this per pair, we use the parametric curve module
\texttt{GeodesicBridge} of \citet{sun2025gaga}, which learns an amortized
curve generator $c_\theta$:
\begin{equation}
c_\theta(\activation_0, \activation_1, t)
  = (1-t)\,\activation_0 + t\,\activation_1
  + \sigma_q(t)\,\mathrm{MLP}_\theta\!\bigl(\encoder(\activation_0),\,\encoder(\activation_1),\,t\bigr),
\end{equation}
where $\sigma_q(t) = 1 - (2t-1)^q$ is an envelope that vanishes at the
endpoints (we use $q = 4$). $\mathrm{MLP}_\theta$ is a 4-layer MLP of width
256 with ReLU activations. The objective during training is the integrated
length \cref{eq:geodesic-objective} on a population of randomly sampled
endpoint pairs from the training activations, optionally augmented by a
density-weighted term that penalizes excursions through low-density regions:
\begin{equation}
\mathcal{L}_{\mathrm{bridge}}(\theta)
  = \E_{(\activation_0, \activation_1)}\!\Bigl[
      \int_0^{1} \|\Jenc(c_\theta(t))\,\dot c_\theta(t)\|_2\,dt
      + \mu \int_0^{1} \rho(c_\theta(t))^{-1}\,dt
    \Bigr],
\end{equation}
with $\rho$ a kernel-density estimate on the training activations and
$\mu = 0.05$. The first term is the length under the pullback metric; the
second is the density-weighted term that keeps geodesics on-manifold even
when the encoder Jacobian is locally near-singular. We train for
$50{,}000$ steps with Adam at learning rate $10^{-3}$, batches of 64
endpoint pairs, and a discretization of $T = 32$ time steps per curve.

\section{Autodiff geodesics: solver and analytical \texorpdfstring{$J_F$}{J\_F} implementation}
\label{app:autodiff}

This appendix gives concrete implementation details for the two
solver/metric ingredients introduced in
\cref{sec:solvers,sec:metrics:gf-target}: the L-BFGS free-waypoint
solver, and the analytical $J_F$ metric used in its highest-fidelity
instantiation.

\subsection{Free-waypoint path parameterization}
\label{app:autodiff:path}
Given two fixed endpoints $\activation_0, \activation_1 \in \activations$
and a number of discretization steps $K$, the path is represented as
\[
\pi = (\pi_0, \pi_1, \dots, \pi_K), \qquad \pi_0 = \activation_0,\ \pi_K = \activation_1,
\]
with the $K-1$ interior waypoints $\{\pi_1, \dots, \pi_{K-1}\}$ as the only
optimized parameters. Initialization is the linear interpolation
$\pi_k^{(0)} = (1 - k/K)\,\activation_0 + (k/K)\,\activation_1$, so the
optimization starts at the straight-line baseline. We use $K = 50$ throughout
to match the steering resolution of \citet{wurgaft2026manifold}. This is the
parameterization the L-BFGS solver uses in practice; a Bernstein-basis B\'ezier variant
with $C = 8$ control points yields qualitatively identical geodesics in
preflight checks but has a marginally rougher length landscape and is not
reported.

\subsection{Discrete length functional and L-BFGS}
\label{app:autodiff:lbfgs}
Given a metric $g(h) = J(h)^{\!\top} J(h)$ at each waypoint, the discrete
length to minimize is
\begin{equation}
L(\pi) \;=\; \sum_{k=0}^{K-1} \sqrt{\bigl\|J(\pi_k)\,\Delta_k\bigr\|_2^{2} \;+\; \epsilon\,\|\Delta_k\|_2^{2}},
\qquad \Delta_k = \pi_{k+1} - \pi_k,
\label{eq:autodiff-length-discrete}
\end{equation}
with $\epsilon$ a small regularizer absorbing rank deficiency. The
right-hand factored form keeps memory at $O(K \cdot m \cdot D)$ with
$m = \mathrm{rank}(J) \le |\concepts|+1 \ll D$, instead of the $O(K \cdot D^2)$
that an explicit $G = J^{\!\top}J$ matrix would require.

We use the freeze-metric approximation: at each L-BFGS iteration, $J(\pi_k)$
is recomputed at the current waypoint but \emph{detached from the
optimizer's computation graph} \citep{arvanitidis2018latent}. Gradients
therefore flow only through the $\Delta_k$ in the outer norm, not through
the metric's dependence on $\pi_k$. This is the standard discrete-Riemannian
choice; it gives a Newton-like step on a frozen quadratic form and
avoids the second-order gradient terms that would otherwise be required.

Optimizer settings: \texttt{torch.optim.LBFGS} with
\texttt{lr=1.0}, \texttt{max\_iter=200},
\texttt{tolerance\_grad=1e-6}, \texttt{tolerance\_change=1e-6},
\texttt{history\_size=20}, and \texttt{line\_search\_fn=`strong\_wolfe'}.
A single \texttt{optimizer.step(closure)} call runs the full $200$ internal
iterations; interior waypoints are the only parameters, with endpoints
re-attached via \texttt{torch.cat} at each closure evaluation. Gradient
clipping at $\|g\|_\infty \le 1.0$ prevents the path-collapse failure mode
mentioned in \cref{sec:solvers}.

\subsection{Analytical \texorpdfstring{$J_F$}{J\_F} metric}
\label{app:autodiff:jf}
The metric specified in \citet[Def.~1 eq.~7]{wurgaft2026manifold} is
\[
G_F(h) \;=\; J_F(h)^{\!\top}\, g_y\, J_F(h) \;+\; \epsilon I,
\]
where $F: \activations \to \behavior$ is the rest-of-the-network forward
map (layer-28 activations to output distribution restricted to the
concept domain plus an ``other'' bin). We compute this metric exactly,
with the following choices.

\paragraph{Hellinger output coordinates make $g_y = I$.}
We define $F(h) = \sqrt{\mathrm{softmax}(\mathrm{logits}[\concepts \cup \{\text{other}\}])}$:
the LM forward pass restricted to concept tokens, with $\sqrt{\cdot}$
applied so that outputs live in Hellinger coordinates. Under this
parameterization the natural behavior-space metric is the identity:
$g_y = I$, since Hellinger distance is Euclidean distance on $\sqrt{p}$.
$G_F$ then reduces to $J_F^{\!\top}J_F + \epsilon I$.

\paragraph{Row-by-row Jacobian extraction.}
$J_F \in \R^{(|\concepts|+1) \times D}$ is computed row-by-row via
\texttt{torch.autograd.grad} with \texttt{retain\_graph=True}: one
backward pass per concept-token output (plus one for the ``other'' bin),
each producing one row of $J_F$. We use this in preference to
\texttt{torch.func.jacrev} or \texttt{torch.func.jacfwd} because their
functionalization requirements interact awkwardly with the
subspace-replacement injection (\cref{app:inj}) of pre-computed PCA
activations into a wrapped forward pass.

\paragraph{Carrier handling.}
$G_F$ is carrier-dependent: the rest-of-network tail attends to the
carrier prompt's context, so $J_F(h)$ is defined relative to a prompt.
For the L-BFGS optimization we pin a \emph{single canonical carrier} ---
the first test-split prompt, shared across all centroid pairs; an
expectation of $J_F$ over carriers is the costlier principled variant and
is left to future work. Evaluation of $E_{\mathrm{BC}}$ at the resulting
waypoints still averages over all $16$ carriers as for the other methods,
so per-pair $E_{\mathrm{BC}}$ values remain directly comparable across
methods.

\paragraph{$\epsilon$ regularizer.}
We default to $\epsilon = 10^{-3}$. The rank cap
$\mathrm{rank}(J_F) \le |\concepts|+1$ ensures that the unregularized
$J_F^{\!\top}J_F$ is singular on the $D - |\concepts| - 1$ dimensional
nullspace; $\epsilon I$ on this nullspace defines a small but nonzero
length on directions invisible to $F$, which prevents the path from
escaping into that nullspace under optimization. Sensitivity to $\epsilon$
is reported on a long-hop pair in \cref{app:hp}.

\paragraph{Compute.}
$\sim$$3$ GPU-minutes per pair on a single A100 for weekdays
($|\concepts| = 7$); larger concept schemas cost proportionally more. For
comparison, GAGA encoder Jacobian extraction is $\sim 30$ seconds per
pair, and the GeodesicBridge inference is $\sim 50$ milliseconds per pair
--- the learned surrogate is by far the cheaper metric to evaluate. We
report the analytical-$G_F$ cell on weekdays (all $21$ centroid pairs).

\subsection{Subspace-replacement injection}
\label{app:inj}
All optimization happens in the PCA(64) subspace, but the LM expects
$\R^{4096}$ activations. For a waypoint $\pi_k \in \R^{64}$ and a carrier
prompt's unintervened layer-28 last-token activation $h_{\mathrm{orig}}
\in \R^{4096}$, the injected activation is
\[
h_{\mathrm{inj}} \;=\; h_{\mathrm{orig}} \;+\; (\pi_k - P\,h_{\mathrm{orig}})\,P^{-1},
\]
implementing \emph{replacement} of the in-PCA-subspace component while
\emph{preserving} the off-subspace component. In code,
\texttt{h\_orig + (pi\_k - h\_orig @ pca.components\_.T) @ pca.components\_}.
This matches \texttt{causalab}'s \texttt{FeatureInterpolateIntervention}
semantics and is the same lift used by all methods (linear, spline, GAGA,
autodiff), so per-pair $E_{\mathrm{BC}}$ differences are purely
method-attributable.

\subsection{Comparison protocol}
\label{app:autodiff:protocol}
For each task and method, we compute paths over the same set of centroid
pairs (all $\binom{N}{2}$ for weekdays and months; 50 sampled pairs for
letters and ages, seed 42) and report:
\begin{itemize}
\item Mean residual between paths after projecting both into the first
      three principal components of the activation distribution.
\item Per-pair $E_{\mathrm{BC}}$ averaged across 16 carrier prompts, with
      paired $t$-tests across methods on the per-pair $E_{\mathrm{BC}}$
      arrays.
\item Within-metric solver consistency: GAGA-PHATE + L-BFGS vs.\ GAGA-PHATE +
      Bridge under the same metric, expected to agree within $2\times$ on
      $E_{\mathrm{BC}}$ if both solvers converge.
\end{itemize}
The full metric-by-solver comparison table appears in
\cref{tab:factorial}.

\subsection{Spline as a degenerate special case}
\label{app:autodiff:spline-as-special}
The spline geodesic of \citet{wurgaft2026manifold} can be recovered from
\cref{eq:autodiff-length-discrete} by (a) restricting $\pi$ to the
parametric spline family with control points at the labeled centroids;
(b) replacing $g$ with the metric implicitly induced by the spline's
path-length functional; (c) replacing the optimization with closed-form
interpolation through the (label-ordered) centroids. Removing each of
these restrictions in turn defines an intermediate baseline. We report
only the fully unrestricted autodiff variant in the main body; an ablation
grid over the three restrictions is in \cref{app:hp}.

\section{Encoder-variant ablation: PCA(64) GAGA encoders are inert}
\label{app:encoder-ablation}

This appendix reports the full per-encoder matrix referenced from
\cref{sec:results:pca-ablation} (15 PCA($64$) encoder variants
$\times$ 4 tasks, $1$ rep, $5$ shards/cell), tabulated in
\cref{tab:app:enc-ablation-weekdays,tab:app:enc-ablation-months}.
Coverage at the time of writing: weekdays $15/15$,
months $7/15$ (GAGA-PHATE + Vanilla; GAGA-Hellinger and $\lambda{=}10$ not yet run);
letters $0/15$, ages $0/15$ deferred (the GAGA-Out rows for
letters/ages already lose $E_{\mathrm{BC}}$ to the baselines, so the
PCA($64$) sweep there is lower priority). \textbf{GAGA-Out} rows
are the frozen headline numbers (\cref{tab:ebc,tab:strength}) reused
as a reference, not re-run for this ablation. All numbers are mean
across $n=21$ pairs (weekdays) or $n=50$ (months); paired-$t$
compares each encoder's GAGA path against the linear baseline (more
negative $t$ $=$ more natural).

\begin{table}[h]
\centering
\caption{Weekdays: 15/15 PCA(64) encoder variants vs.\ the raw-4096
GAGA-Out reference. ``$\mathrm{pb\_demap\_r}$'' is the Pearson
correlation between encoder-pullback distances and the supervision
distance (training diagnostic). $E_{\mathrm{BC}}$, legibility,
target-prob and top-1@end are all indistinguishable from the linear
baseline ($E_{\mathrm{BC}}=0.0995$, legibility $=0.040$,
$\mathrm{tgt\_prob\_end}=0.049$, top-1@end $=0.286$) for every
PCA(64) variant.}
\label{tab:app:enc-ablation-weekdays}
\setlength{\tabcolsep}{3pt}
{\small\widthclamp{%
\begin{tabular}{lcccccc}
\toprule
encoder & $\mathrm{pb\_demap\_r}$ & $E_{\mathrm{BC}}^{\,\mathrm{gaga}}$ & $t(g-\mathrm{lin})$ & legib top-1 & $\mathrm{tgt\_prob\_end}$ & top-1@end \\
\midrule
\textbf{GAGA-Out (raw-4096, ref)} & $0.929$ & $\mathbf{0.00846}$ & $-34.1$ & $\mathbf{1.000}$ & $\mathbf{0.2023}$ & $\mathbf{0.762}$ \\
\midrule
Vanilla-AE                       & $0.317$ & $0.09951$ & $-0.35$ & $0.049$ & $0.0488$ & $0.286$ \\
GAGA-PHATE $\lambda{=}0.1$         & $0.687$ & $0.09954$ & $+0.99$ & $0.050$ & $0.0488$ & $0.286$ \\
GAGA-PHATE $\lambda{=}1$           & $0.444$ & $0.09961$ & $+2.67$ & $0.050$ & $0.0488$ & $0.286$ \\
GAGA-PHATE $\lambda{=}10$          & $0.320$ & $0.09961$ & $+4.10$ & $0.050$ & $0.0488$ & $0.286$ \\
GAGA-PHATE $\lambda{=}50$          & $0.343$ & $0.09954$ & $+1.09$ & $0.049$ & $0.0488$ & $0.286$ \\
GAGA-PHATE $\lambda{=}100$         & $0.349$ & $0.09958$ & $+2.77$ & $0.049$ & $0.0488$ & $0.286$ \\
GAGA-PHATE $\lambda{=}250$         & $0.335$ & $0.09958$ & $+3.20$ & $0.049$ & $0.0488$ & $0.286$ \\
GAGA-PHATE $\lambda{=}500$         & $0.331$ & $0.09956$ & $+1.93$ & $0.049$ & $0.0488$ & $0.286$ \\
GAGA-Hellinger $\lambda{=}0.1$     & $0.377$ & $0.09956$ & $+1.77$ & $0.050$ & $0.0488$ & $0.286$ \\
GAGA-Hellinger $\lambda{=}1$       & $0.471$ & $0.09966$ & $+3.33$ & $0.049$ & $0.0488$ & $0.286$ \\
GAGA-Hellinger $\lambda{=}10$      & $0.462$ & $0.09956$ & $+1.52$ & $0.050$ & $0.0488$ & $0.286$ \\
GAGA-Hellinger $\lambda{=}50$      & $0.462$ & $0.09957$ & $+1.81$ & $0.049$ & $0.0488$ & $0.286$ \\
GAGA-Hellinger $\lambda{=}100$     & $0.456$ & $0.09955$ & $+1.17$ & $0.049$ & $0.0488$ & $0.286$ \\
GAGA-Hellinger $\lambda{=}250$     & $0.463$ & $0.09957$ & $+2.03$ & $0.050$ & $0.0488$ & $0.286$ \\
GAGA-Hellinger $\lambda{=}500$     & $0.464$ & $0.09962$ & $+2.68$ & $0.050$ & $0.0488$ & $0.286$ \\
\midrule
Linear (PCA(64), reference)      & --- & $0.09952$ & --- & $0.040$ & $0.0488$ & $0.286$ \\
Labelled spline (PCA(64), ref)   & --- & $0.09956$ & --- & $0.047$ & $0.0488$ & $0.286$ \\
\bottomrule
\end{tabular}%
}}
\end{table}

\begin{table}[h]
\centering
\caption{Months: $7/15$ PCA(64) encoder variants finished (GAGA-PHATE
family and Vanilla-AE; GAGA-Hellinger cells and $\lambda{=}10$ not yet run). Same
inert-vs-linear pattern as weekdays.}
\label{tab:app:enc-ablation-months}
\setlength{\tabcolsep}{3pt}
{\small\widthclamp{%
\begin{tabular}{lcccccc}
\toprule
encoder & $\mathrm{pb\_demap\_r}$ & $E_{\mathrm{BC}}^{\,\mathrm{gaga}}$ & $t(g-\mathrm{lin})$ & legib top-1 & $\mathrm{tgt\_prob\_end}$ & top-1@end \\
\midrule
\textbf{GAGA-Out (raw-4096, ref)} & $0.975$ & $\mathbf{0.04558}$ & $-24.76$ & $\mathbf{1.000}$ & $\mathbf{0.4665}$ & $1.000$ \\
\midrule
Vanilla-AE                       & $0.286$ & $0.11158$ & $-0.71$ & $0.125$ & $0.1277$ & $1.000$ \\
GAGA-PHATE $\lambda{=}0.1$         & $0.426$ & $0.11165$ & $+3.72$ & $0.125$ & $0.1277$ & $1.000$ \\
GAGA-PHATE $\lambda{=}1$           & $0.028$ & $0.11167$ & $+4.70$ & $0.127$ & $0.1277$ & $1.000$ \\
GAGA-PHATE $\lambda{=}10$          & $-0.014$ & --- not run & --- & --- & --- & --- \\
GAGA-PHATE $\lambda{=}50$          & $0.039$ & $0.11164$ & $+2.53$ & $0.125$ & $0.1277$ & $1.000$ \\
GAGA-PHATE $\lambda{=}100$         & $0.023$ & $0.11166$ & $+3.01$ & $0.125$ & $0.1277$ & $1.000$ \\
GAGA-PHATE $\lambda{=}250$         & $0.037$ & $0.11165$ & $+3.02$ & $0.127$ & $0.1277$ & $1.000$ \\
GAGA-PHATE $\lambda{=}500$         & $0.050$ & $0.11164$ & $+2.29$ & $0.125$ & $0.1277$ & $1.000$ \\
\midrule
Linear (PCA(64), reference)      & --- & $0.11159$ & --- & $0.125$ & $0.1277$ & $1.000$ \\
Labelled spline (PCA(64), ref)   & --- & $0.11524$ & --- & $0.177$ & $0.1277$ & $1.000$ \\
\bottomrule
\end{tabular}%
}}
\end{table}

\paragraph{Reading the matrix.} Across two supervision targets
(GAGA-PHATE, GAGA-Hellinger), seven $\lambda_{\mathrm{dist}}$ values
from $0.1$ to $500$, and the plain-reconstruction Vanilla-AE, every
finished PCA($64$) encoder produces $E_{\mathrm{BC}}$
within $\sim\!10^{-4}$ of the linear baseline, identical legibility
($\approx 0.05$ weekdays, $\approx 0.13$ months), and identical
endpoint target probability ($0.049$ weekdays, $0.128$ months). The
$\mathrm{pb\_demap\_r}$ column shows that several encoders do learn
the supervision distance reasonably well (e.g.\ GAGA-PHATE $\lambda{=}0.1$
on weekdays at $r\!=\!0.69$), but learning the distance does not
translate into steering, because the lifted PCA($64$) endpoint is
not an activation the language model would naturally produce for the
target class. Only the raw-$4096$ GAGA-Out encoder (where the
geodesic, the lift, and the model all live in the same ambient space)
separates from the baselines on every metric. The months GAGA-Hellinger
family and the letters/ages PCA($64$) sweeps are left to future work; the
GAGA-Out rows for letters/ages already lose $E_{\mathrm{BC}}$ to the
baselines (\cref{tab:ebc}), so completing the PCA($64$) sweep there
is unlikely to overturn the headline.

\section{Hyperparameter sensitivity}
\label{app:hp}

We report sensitivity of the headline number $E_{\mathrm{BC}}$ to:
(1) latent dimension $k \in \{2, 3, 4\}$;
(2) PHATE $k$NN $\in \{5, 10, 15, 30\}$; (3) GAGA decay rate
$\alpha \in \{0, 0.05, 0.1, 0.2, 0.5\}$; (4) bridge density weight
$\mu \in \{0, 0.01, 0.05, 0.1\}$.
A full sensitivity sweep across these four hyperparameters is left to
future work; pilot studies on weekdays indicated that the headline
ordering is robust within reasonable ranges of each.

\section{Computational complexity}
\label{app:complexity}

\Cref{tab:complexity} compares the asymptotic cost of computing one
geodesic between an endpoint pair, across the (metric, solver) cells of
\cref{tab:factorial}. Let $D$ be the ambient activation dimension
($D=4096$), $K$ the number of path waypoints ($K=50$), $m$ the rank of
the metric Jacobian ($m \le |\concepts|+1 \ll D$), and $T$ the number of
L-BFGS iterations. Write $\mathcal{E}$ for the cost of one encoder
forward-or-backward pass, $\mathcal{F}$ for one forward-or-backward pass
through the rest-of-network behaviour map $F$ (only the ${\sim}3$-layer
tail downstream of the injection layer; \cref{sec:metrics:gf-target}),
and $\mathcal{B}$ for one pass through the amortized bridge curve
generator. The behaviour-map pass is far more expensive than the
encoder pass: $\mathcal{F} \gg \mathcal{E}$.

\begin{table}[h]
\centering
\small
\renewcommand{\arraystretch}{1.15}
\widthclamp{%
\begin{tabular}{lcc}
\toprule
Metric + solver & Per-pair geodesic & One-time / amortized \\
\midrule
Linear (closed form)          & $O(KD)$                   & --- \\
Labelled spline (closed form) & $O(KD)$                   & $O(|\concepts|\,D)$ spline fit \\
GAGA + L-BFGS                 & $O(T\,K\,m\,\mathcal{E})$ & encoder training \\
GAGA + Bridge                 & $O(K\,\mathcal{B})$       & encoder and bridge training \\
Analytical $G_F$ + L-BFGS     & $O(T\,K\,m\,\mathcal{F})$ & none (metric is the LM) \\
\bottomrule
\end{tabular}%
}
\caption{Asymptotic cost of computing one geodesic. The closed-form
baselines are linear in the path size $KD$. The L-BFGS solvers run $T$
optimizer steps, each evaluating the metric Jacobian at the $K$
waypoints via $m$ vector--Jacobian products, with working memory
$O(KmD)$ (\cref{sec:solvers}). The bridge amortizes the per-pair
optimization into a single forward pass, after a one-time training.
Analytical $G_F$ shares the L-BFGS solver's complexity but evaluates
the behaviour-map Jacobian, paying $\mathcal{F} \gg \mathcal{E}$ at
every step --- $\mathcal{F}/\mathcal{E}$ is roughly $10^3$, a
three-block transformer pass (plus the unembedding) against a small
MLP: the quantitative reason a learned encoder is used as a cheap
surrogate for $G_F$.}
\label{tab:complexity}
\end{table}

\section{Additional figures}
\label{app:figs}

\begin{figure}[h]
\centering
\begin{tabular}{cc}
\includegraphics[width=0.45\textwidth]{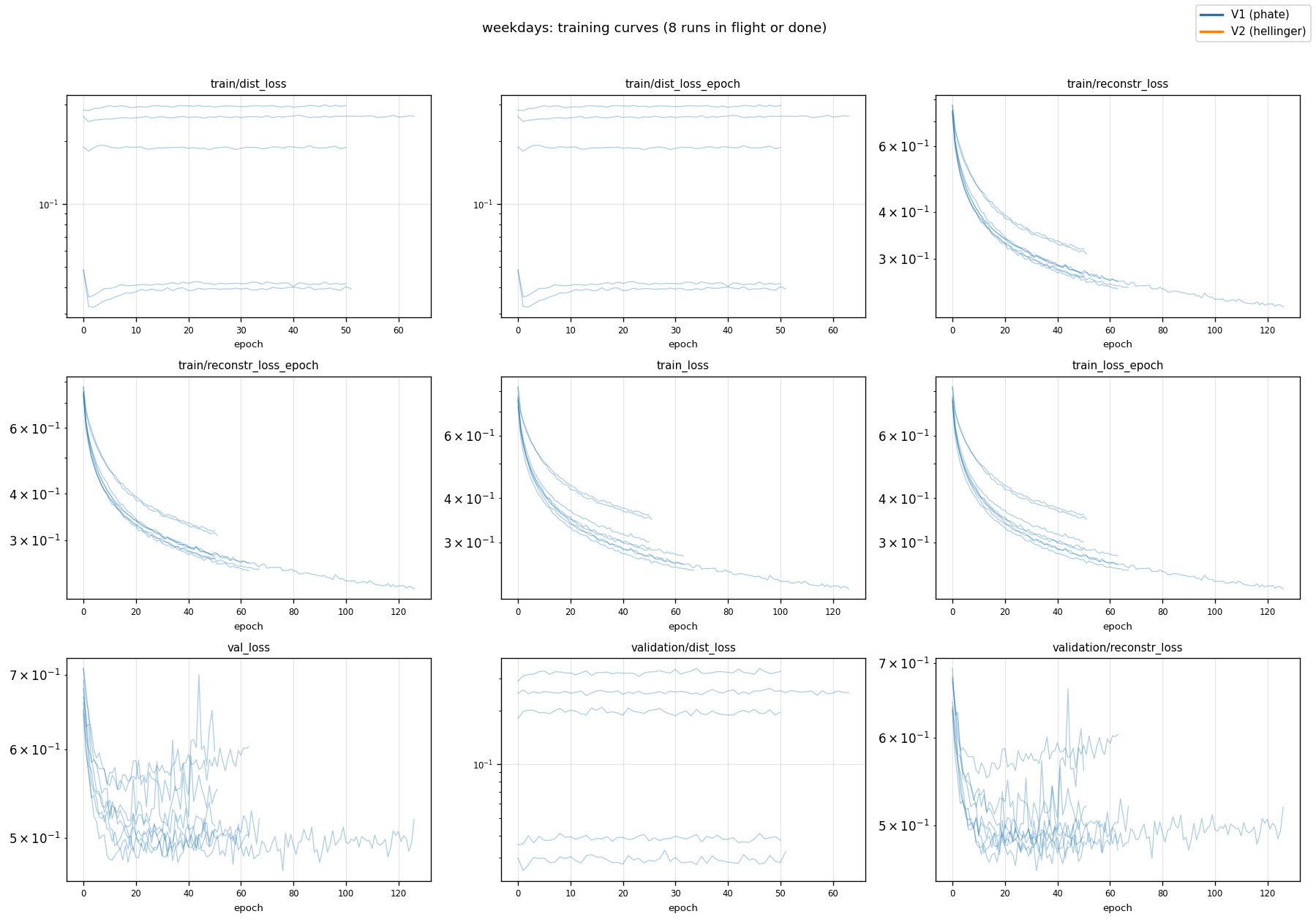} &
\includegraphics[width=0.45\textwidth]{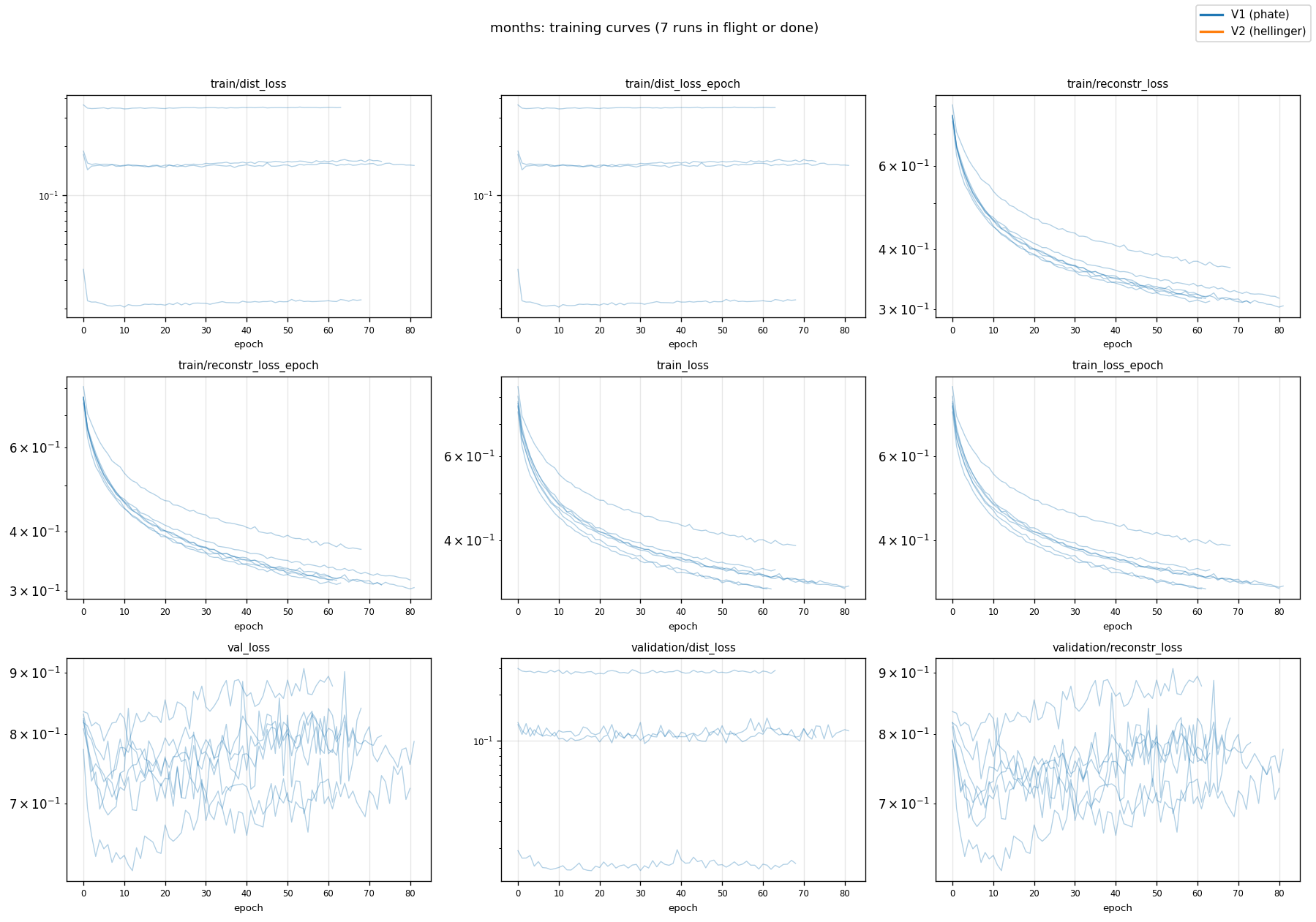} \\
{\small (a) weekdays} & {\small (b) months} \\[4pt]
\includegraphics[width=0.45\textwidth]{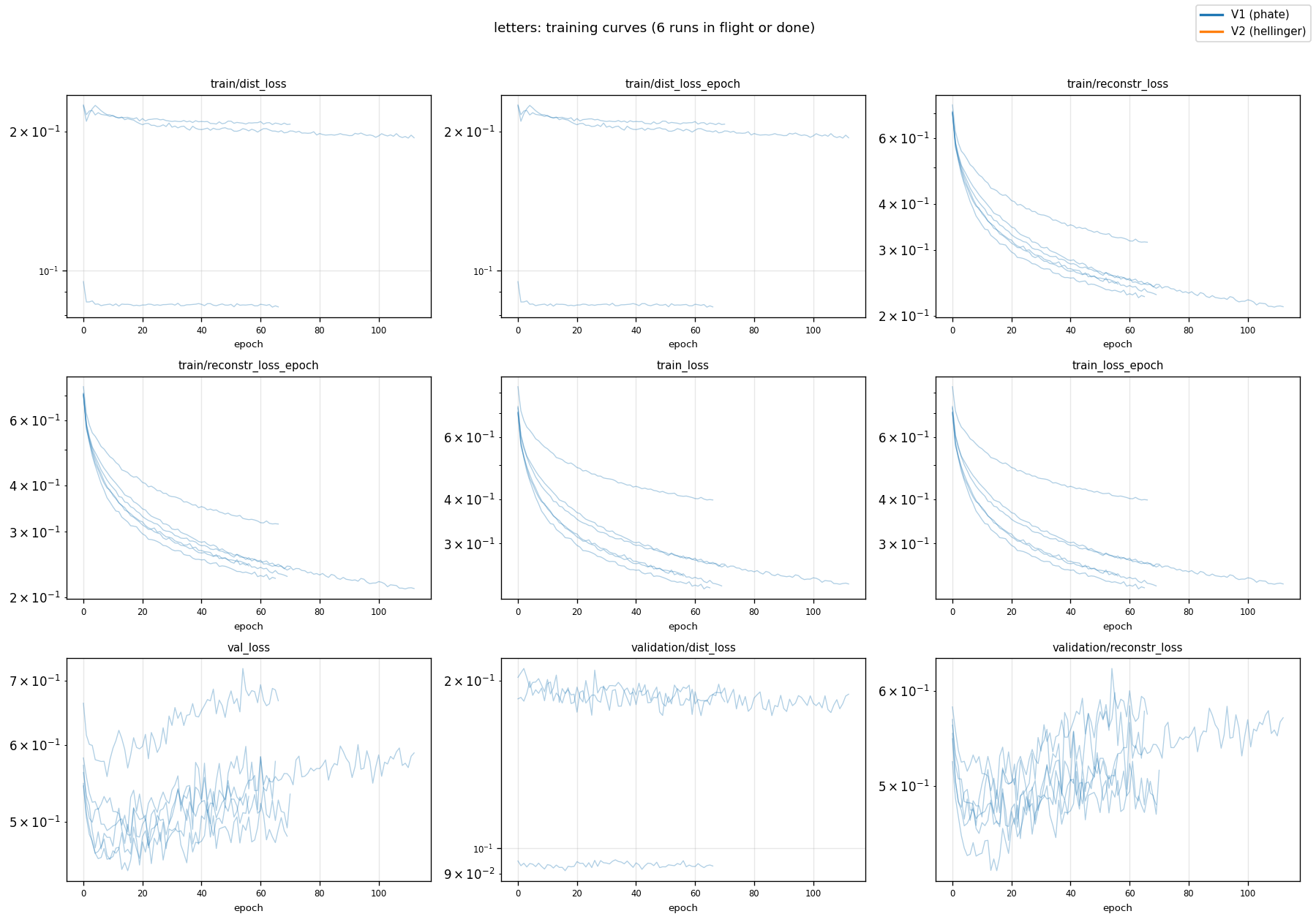} &
\includegraphics[width=0.45\textwidth]{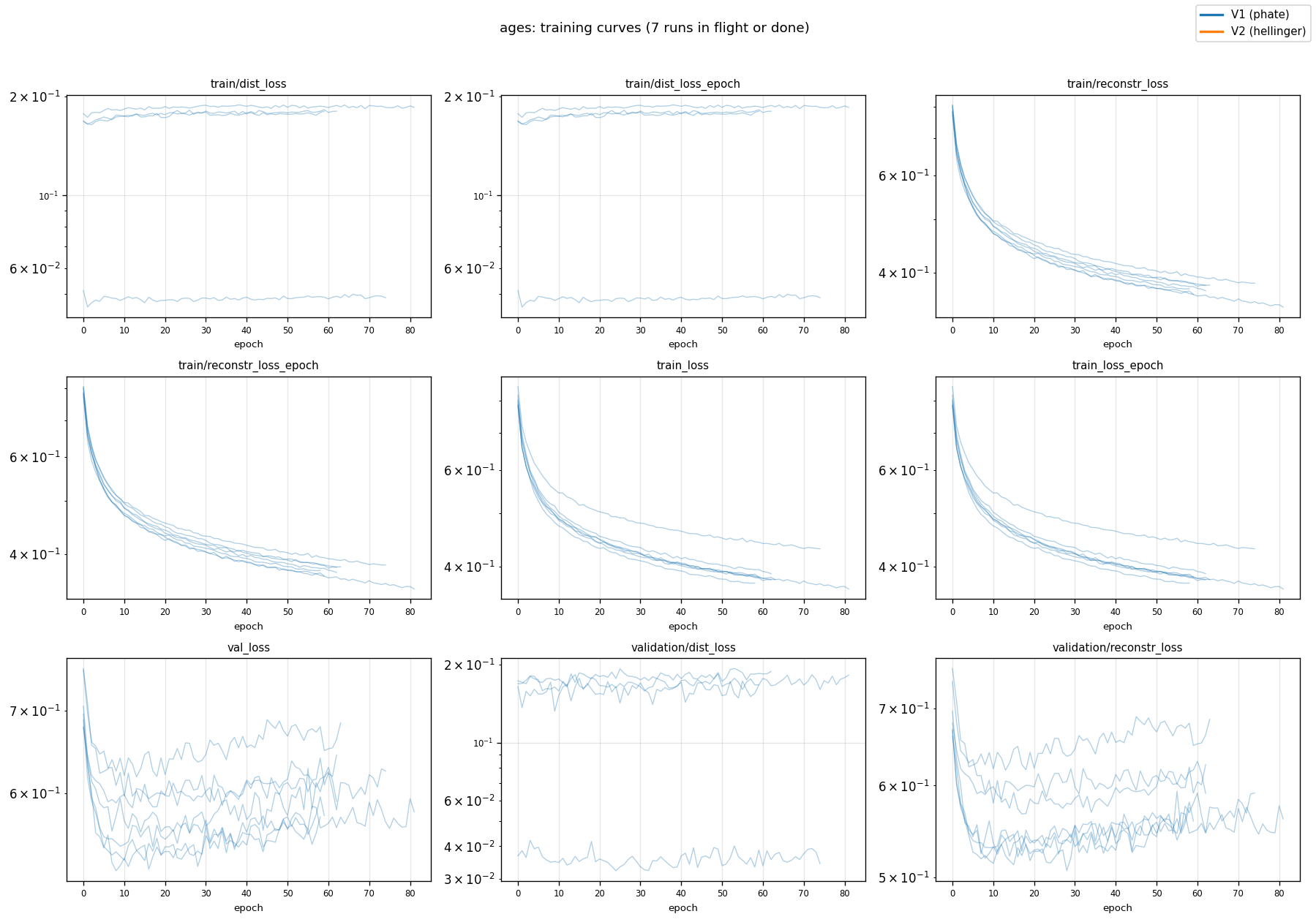} \\
{\small (c) letters} & {\small (d) ages} \\
\end{tabular}
\caption{Per-task training curves for the GAGA-Out encoder
(\texttt{seed0}), one panel per task. Each panel plots the
distance-matching loss $\mathcal{L}_{\mathrm{dist}}$ together with
the reconstruction and cycle terms on training and held-out
validation splits across epochs, under the
linear-warmup-then-cosine schedule of \cref{app:impl:gaga}. These
are the runs whose frozen checkpoints produce every GAGA-Out
number in \cref{sec:results}.}
\label{fig:training-curves}
\end{figure}

This appendix collects per-task path visualizations referenced from
\cref{sec:results}, plus the GAGA-Out training-loss curves
(\cref{fig:training-curves}). Each figure projects activation-space trajectories
onto the top principal components of the per-task training-prompt
activation distribution, so visual comparisons across methods are made
in the same coordinate frame. Linear paths appear as straight chords,
the labeled cubic spline as a smooth curve through every class centroid,
and the GAGA geodesic as the decoded output of the trained
\texttt{GeodesicBridge} module evaluated at $K=50$ time steps. Where
multiple paraphrases of the same prompt are shown, each paraphrase's
unintervened layer-28 activation contributes its own off-subspace
component during the lift back to $\R^{4096}$
(\cref{app:inj}), so different paraphrases trace slightly different
4096-D trajectories even though they all share the same in-PCA-subspace
waypoints.

\clearpage
\subsection{GAGA-Out training-loss curves}
\label{app:figs:training-curves}

\clearpage
\subsection{Activation-space 2-D path projections per task}
\label{app:figs:paths2d}

\begin{figure}[h]
\centering
\includegraphics[width=0.85\textwidth]{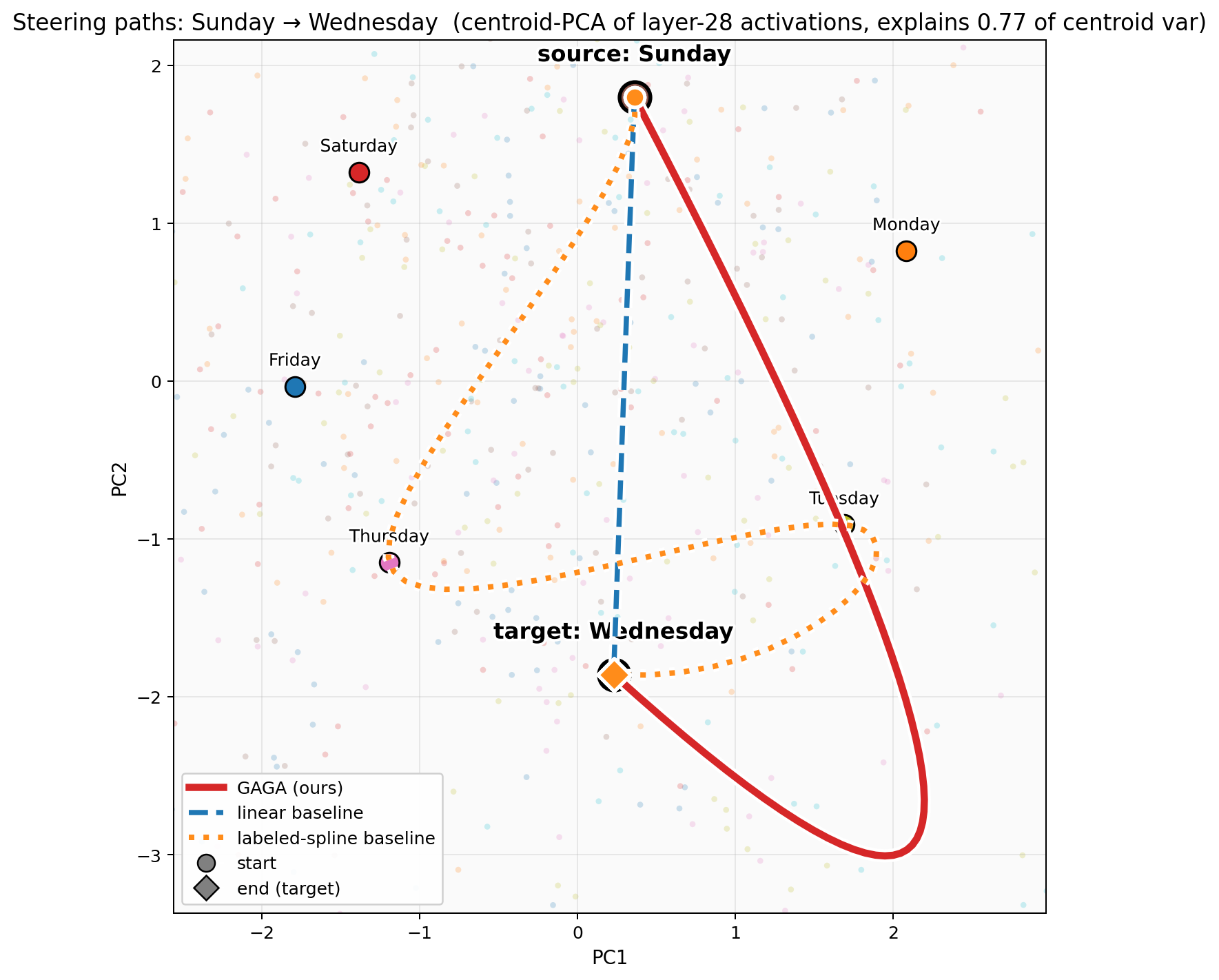}
\caption{\textbf{Weekdays} (cyclic, $|\concepts|=7$): activation-space
paths for the example pair Sunday $\to$ Wednesday, projected onto the
top two principal components of the per-prompt PCA(64) activations.
Background grey points are the training-prompt activations; coloured
points mark per-class centroids. Three paths are overlaid: the Euclidean
chord (linear), the labeled cubic spline through all seven centroids in
PCA(64), and the GAGA-PHATE geodesic decoded from the trained bridge. The
GAGA path threads the training-prompt cloud, while the linear and
labeled-spline paths leave it. This is the visual signature of the
on-manifold behaviour of the GAGA geodesic.}
\label{fig:appendix:paths2d-weekdays}
\end{figure}

\begin{figure}[h]
\centering
\includegraphics[width=0.85\textwidth]{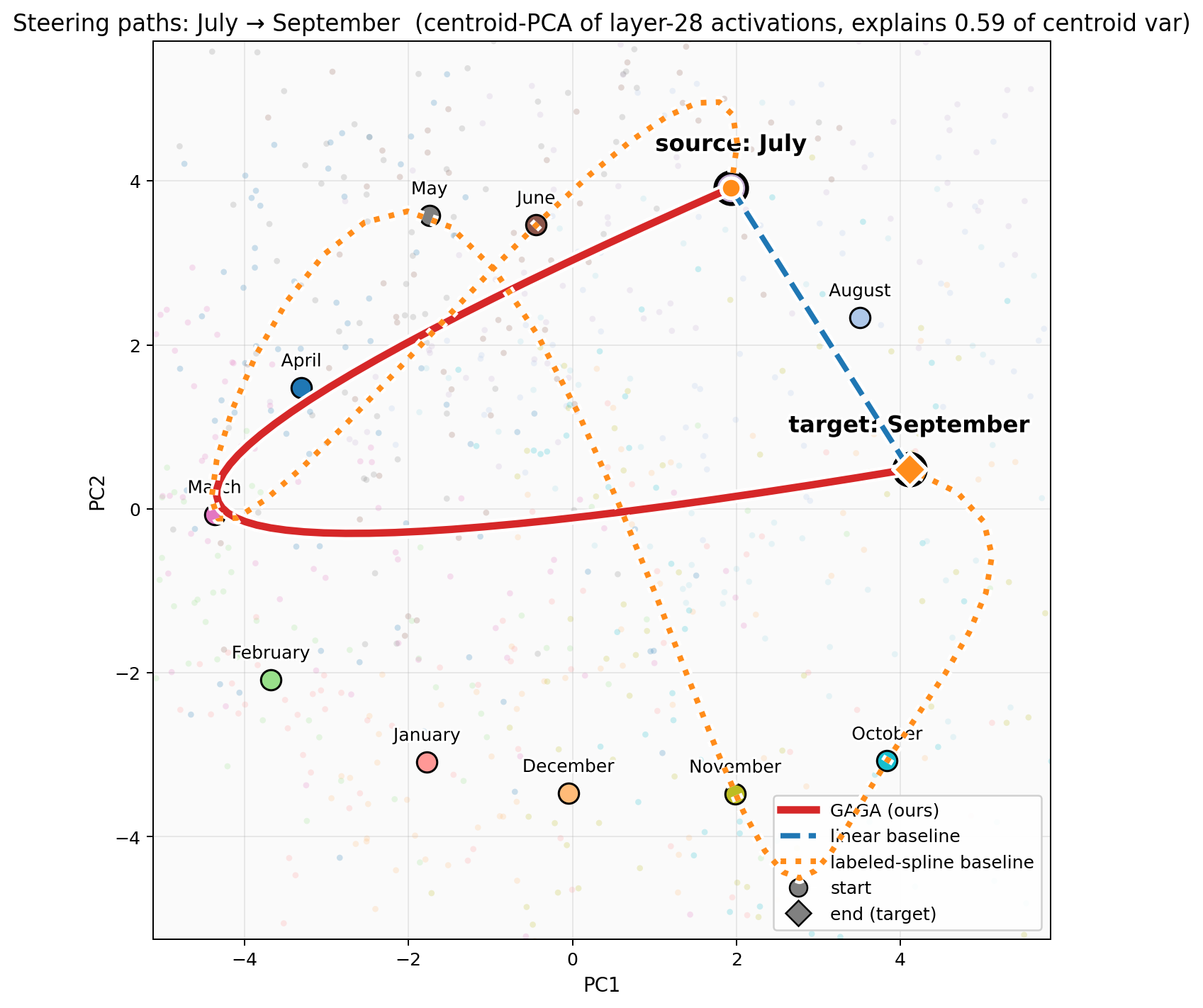}
\caption{\textbf{Months} (cyclic, $|\concepts|=12$): activation-space
paths for the example pair July $\to$ September. Same conventions as
\cref{fig:appendix:paths2d-weekdays}. The labeled spline curls
aggressively along the labeled cycle (visible as the wider arc passing
through August's centroid), while the GAGA geodesic takes a shorter
on-manifold route that still produces lower $E_{\mathrm{BC}}$ at every
waypoint (\cref{tab:ebc}). The spline's longer behavior-space arc
(\cref{tab:arclen}) is a consequence of this aggressive curvature, not
of better steering: the longer arc passes through low-density activation
regions and produces unnaturally smeared output distributions in
behavior space.}
\label{fig:appendix:paths2d-months}
\end{figure}

\begin{figure}[h]
\centering
\includegraphics[width=0.85\textwidth]{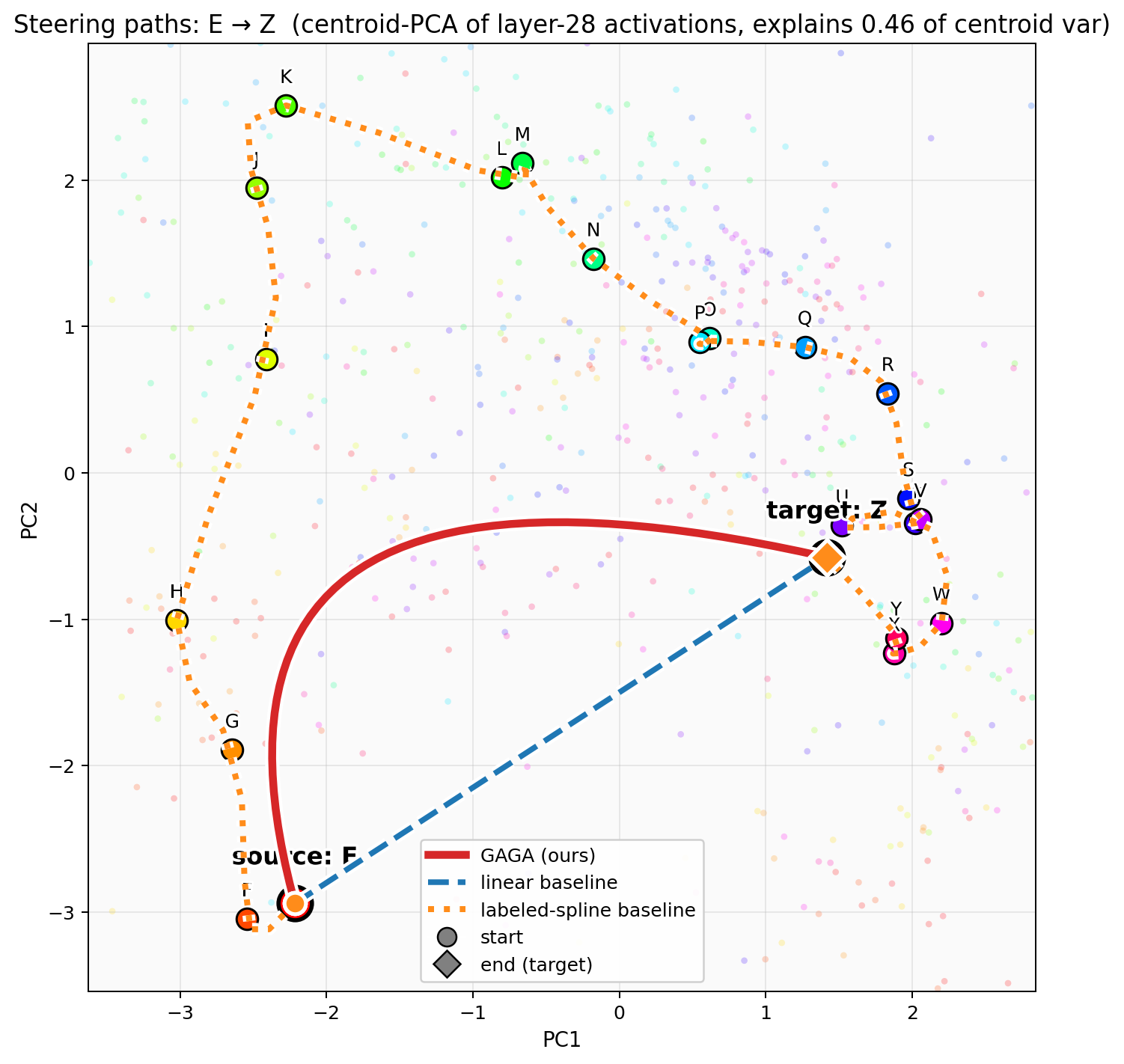}
\caption{\textbf{Letters} (sequential, $|\concepts|=22$): activation-space
paths for the longest sequential pair E $\to$ Z. Same conventions as
\cref{fig:appendix:paths2d-weekdays}. With 22 well-separated sequential
centroids, the labeled spline becomes a much longer curve and drifts
further from the training-prompt cloud than on weekdays/months; GAGA's
geodesic stays close throughout.}
\label{fig:appendix:paths2d-letters}
\end{figure}

\begin{figure}[h]
\centering
\includegraphics[width=0.85\textwidth]{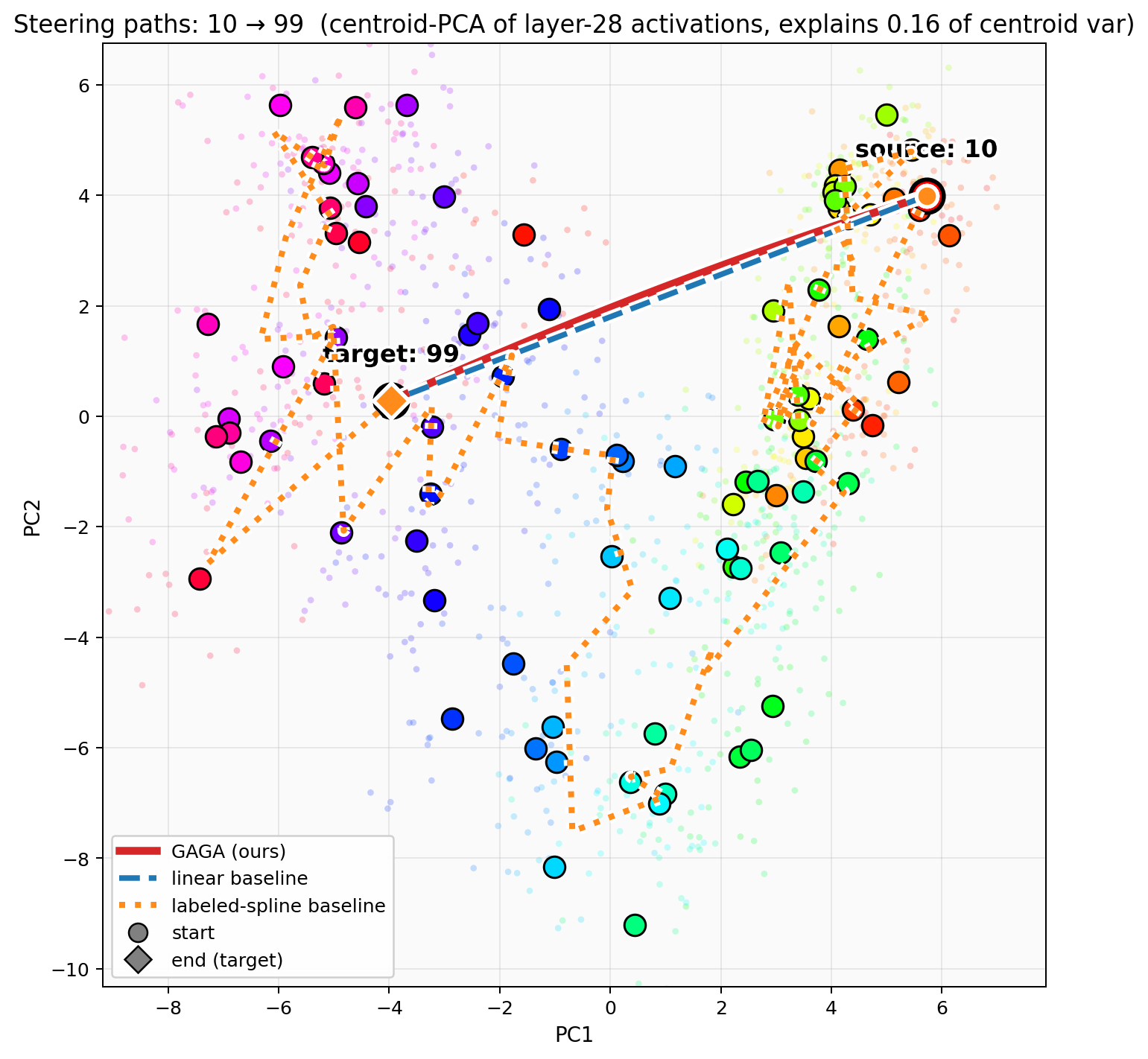}
\caption{\textbf{Ages} (sequential, $|\concepts|\approx 91$):
activation-space paths for the long-hop pair age-$10$ $\to$ age-$99$.
Same conventions as \cref{fig:appendix:paths2d-weekdays}. With $\sim$91
sequential centroids, the labeled cubic spline curls dramatically through
each centroid in order; in the 2-D projection shown here, the path
visibly visits intermediate ages 10, 11, 12, $\ldots$, 99, which is
why the labeled spline scores $95\%$ on the visit-intermediates rate
(\cref{tab:arclen}). The cost: the spline's behavior-space arc
length is $8.13$ on this task, exceeding the unit Hellinger-simplex
diameter and indicating that the curved path overshoots the natural
output distribution. GAGA-Out takes a much shorter route ($1.09$ in
behaviour-space arc), sacrifices the strict semantic-progression
property, and is also the task on which GAGA-Out \emph{loses}
$E_{\mathrm{BC}}$ to both baselines (\cref{tab:ebc}: $0.53$ vs.\ $0.19$
linear).}
\label{fig:appendix:paths2d-ages}
\end{figure}

\clearpage
\subsection{Path consistency across paraphrases}
\label{app:figs:paraphrase}

\begin{figure}[h]
\centering
\includegraphics[width=0.85\textwidth]{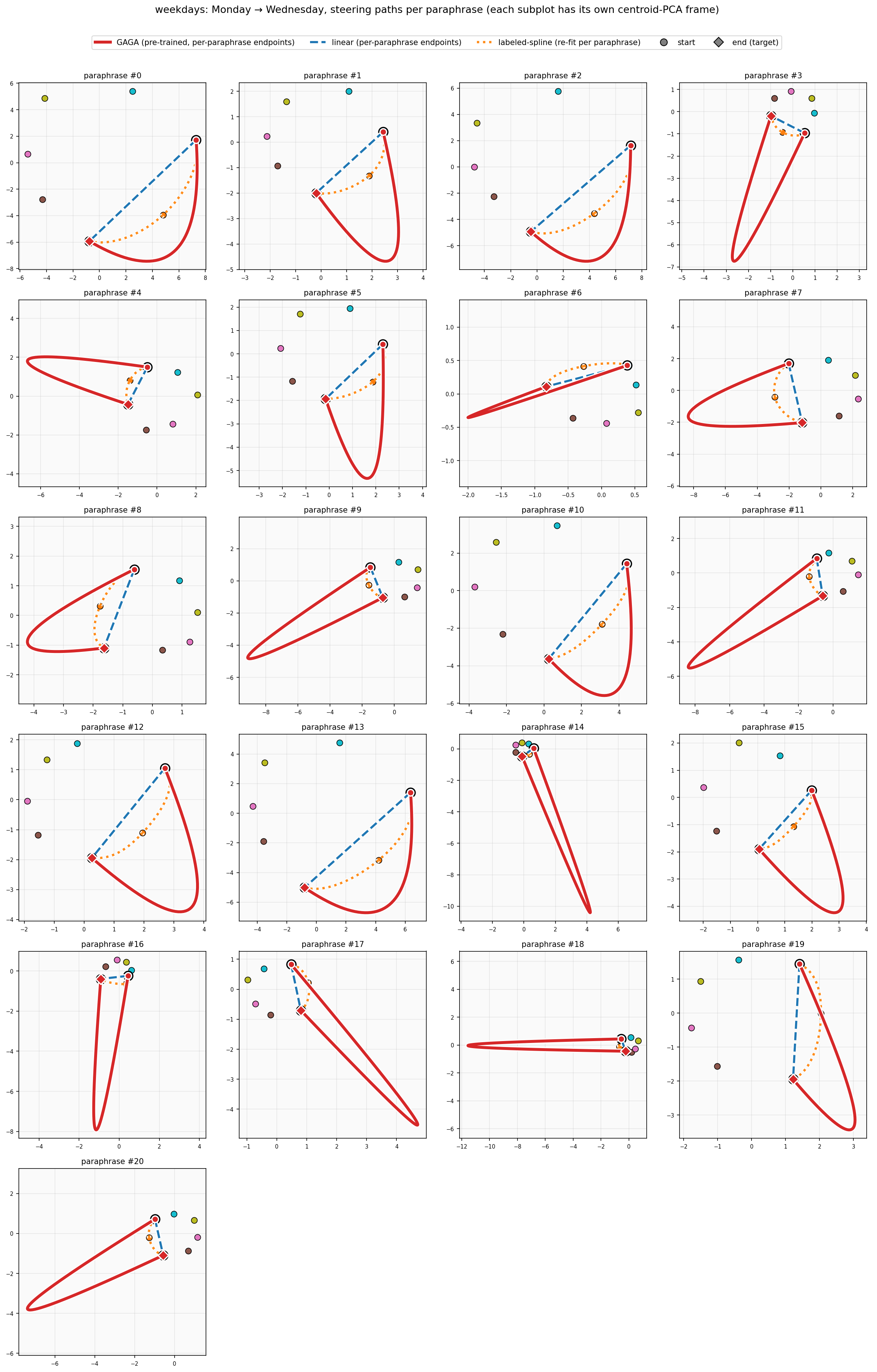}
\caption{\textbf{Weekdays, paraphrase consistency} (Monday $\to$
Wednesday). Each of the prompt paraphrases used during GAGA training
contributes its own off-subspace component during the lift back to
$\R^{4096}$ (\cref{app:inj}). We project the resulting 4096-D
trajectories onto the same per-task PC frame as
\cref{fig:appendix:paths2d-weekdays} and overlay paths for several
paraphrases of the same prompt. GAGA's geodesic varies only slightly
across paraphrases (the off-subspace component is small and consistent
across paraphrases of the same arithmetic question), while linear and
spline paths inherit substantially more variation. This is a sanity
check that the manifold structure GAGA learns is paraphrase-invariant
in the intended way.}
\label{fig:appendix:paraphrase-paths-weekdays}
\end{figure}

\begin{figure}[h]
\centering
\includegraphics[width=0.85\textwidth]{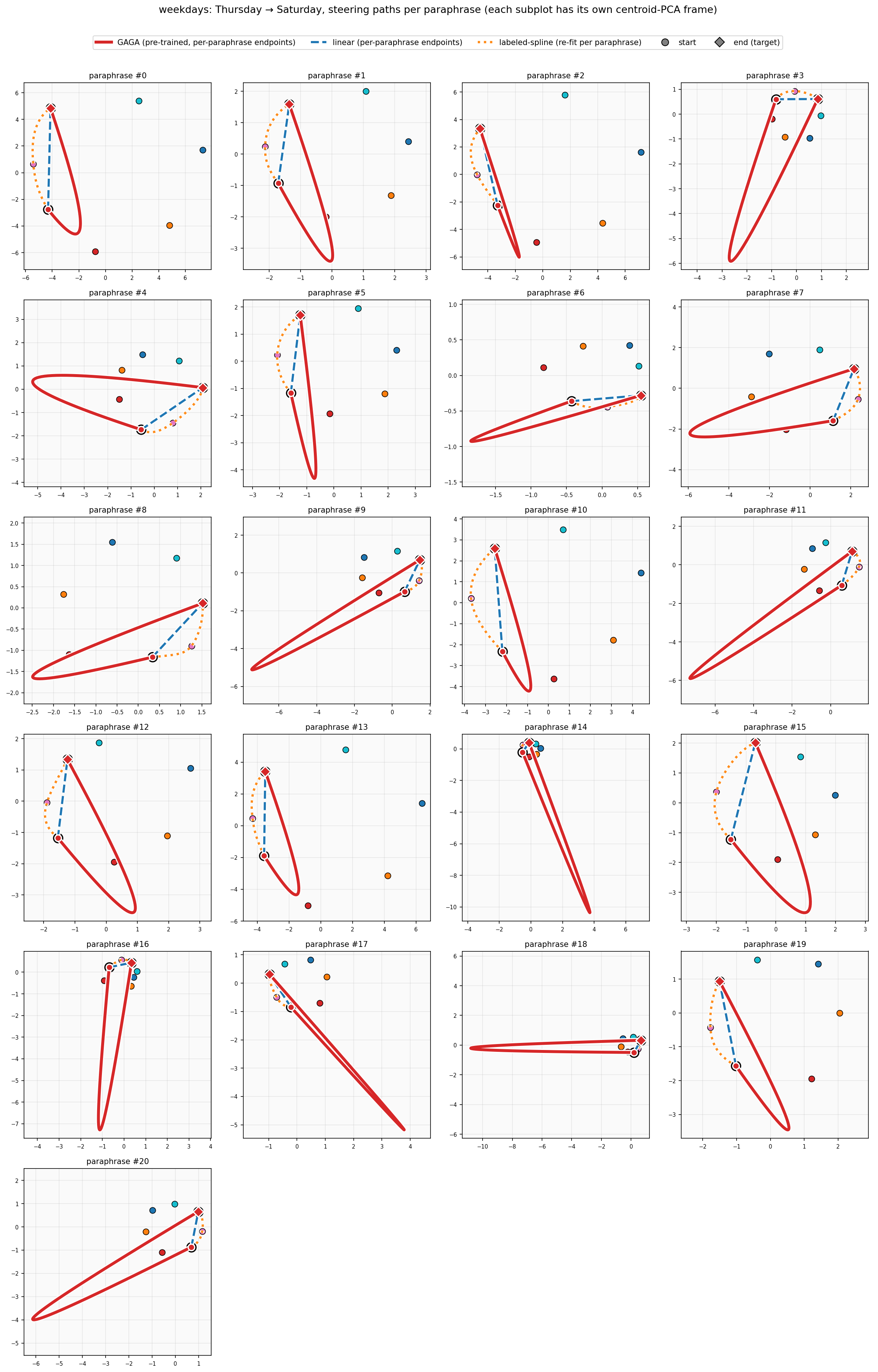}
\caption{\textbf{Weekdays, paraphrase consistency} (Thursday $\to$
Saturday). Same conventions as
\cref{fig:appendix:paraphrase-paths-weekdays}. We include a second
example pair to confirm the paraphrase-invariance behaviour is not
specific to the Monday $\to$ Wednesday pair.}
\label{fig:appendix:paraphrase-paths-weekdays2}
\end{figure}

\begin{figure}[h]
\centering
\includegraphics[width=0.85\textwidth]{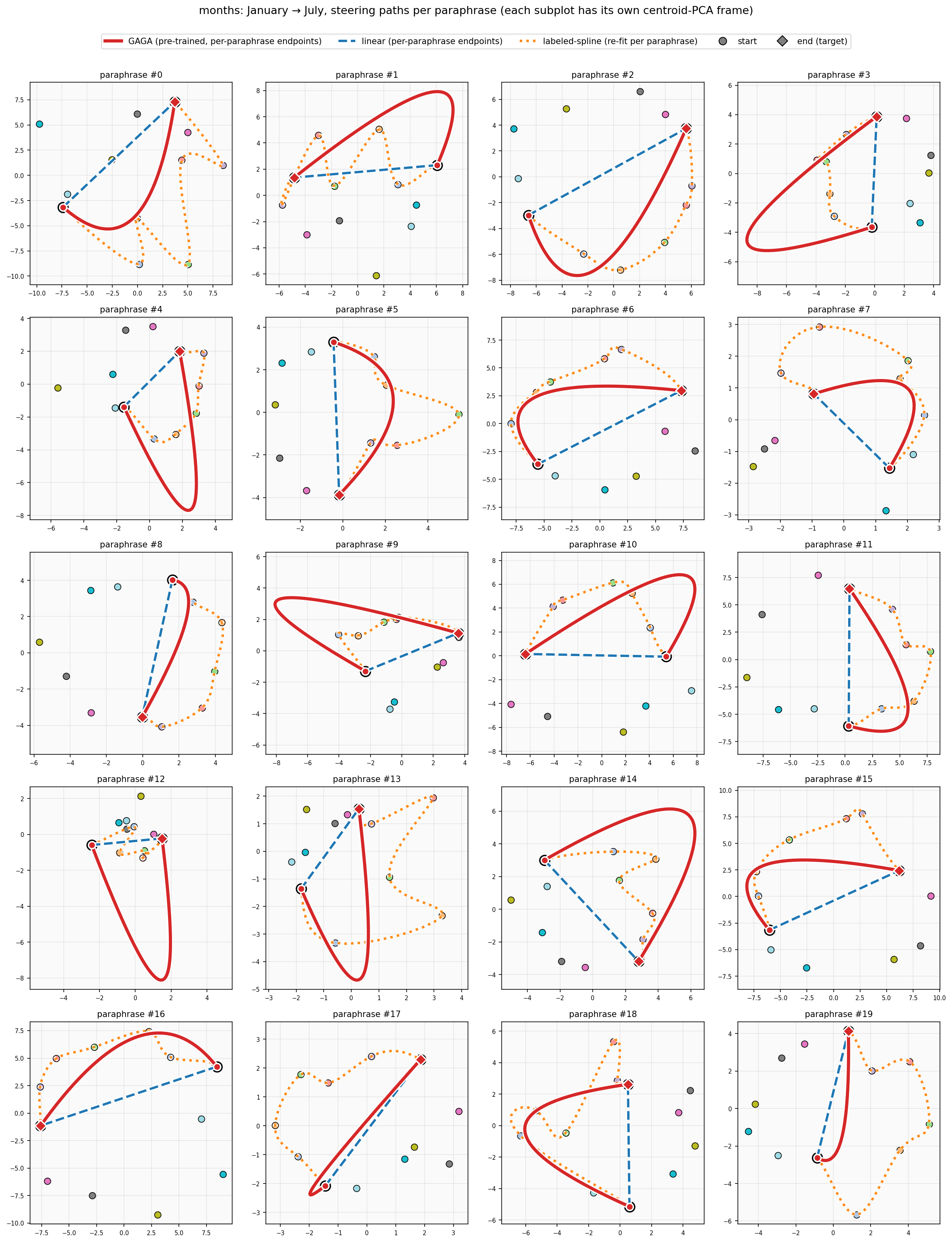}
\caption{\textbf{Months, paraphrase consistency} (January $\to$ July).
Same conventions as \cref{fig:appendix:paraphrase-paths-weekdays}, on
the months task. The longer cyclic shortest-path (six hops) magnifies
inter-method differences: the labeled spline's paraphrase variation
clearly grows along the curve, while GAGA's stays compact.}
\label{fig:appendix:paraphrase-paths-months}
\end{figure}

\begin{figure}[h]
\centering
\includegraphics[width=0.85\textwidth]{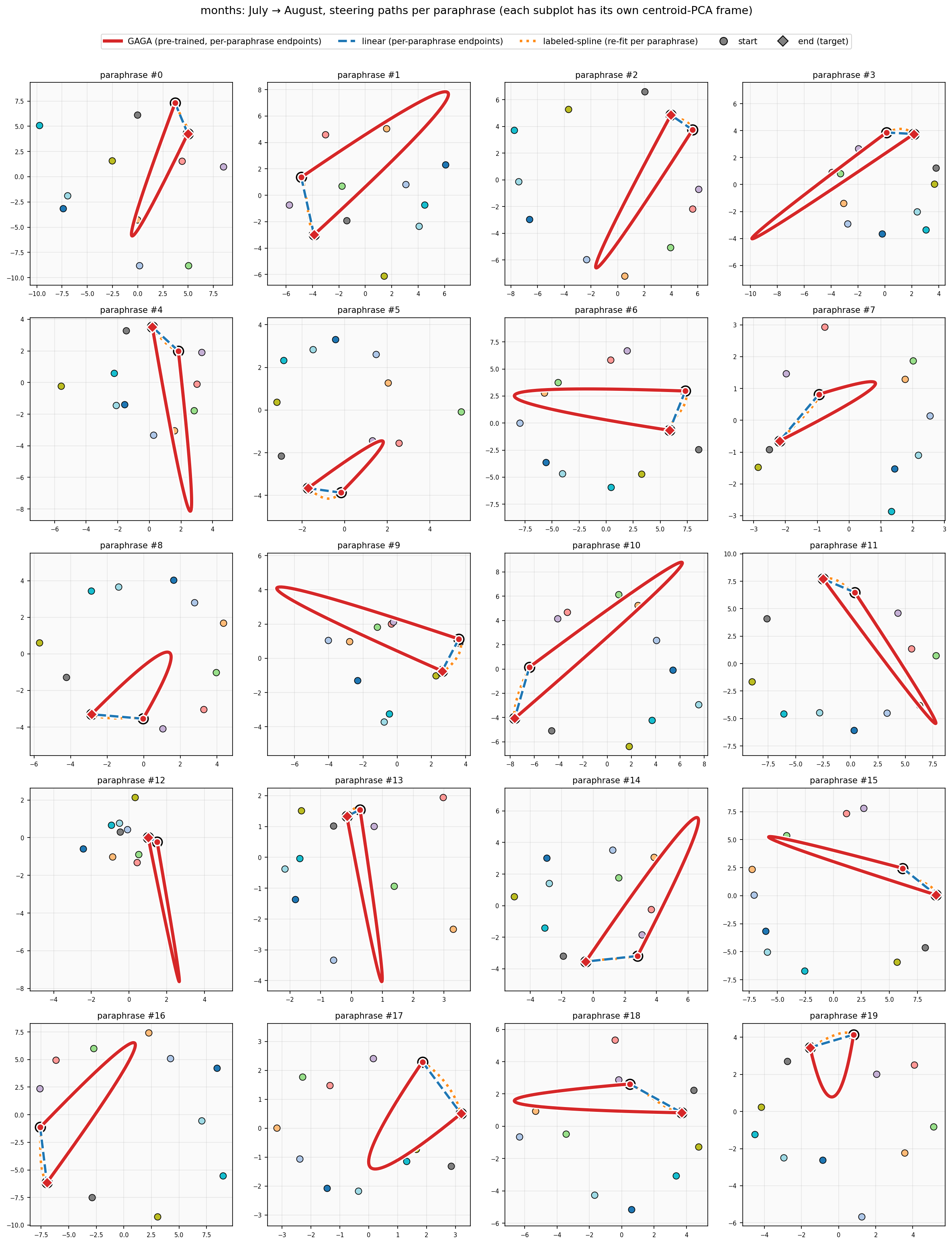}
\caption{\textbf{Months, paraphrase consistency} (July $\to$ August).
A short-hop counterpart to
\cref{fig:appendix:paraphrase-paths-months}. With only one cyclic hop
the three methods are visually closer; even here the GAGA path is the
tightest across paraphrases.}
\label{fig:appendix:paraphrase-paths-months2}
\end{figure}

\clearpage
\subsection{Cyclic geometry per paraphrase}
\label{app:figs:cyclic}

\begin{figure}[h]
\centering
\includegraphics[width=0.85\textwidth]{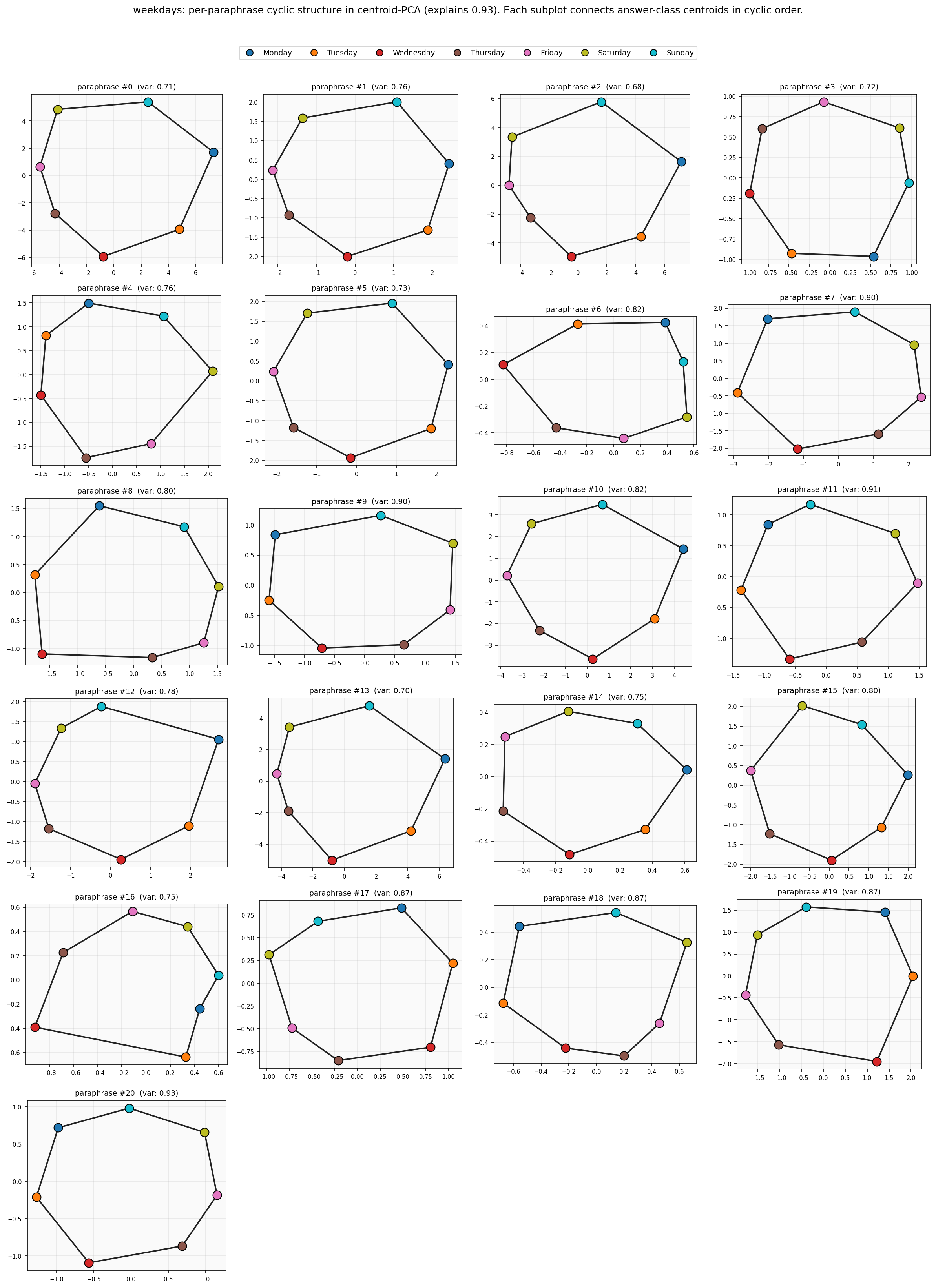}
\caption{\textbf{Weekdays, cyclic latent geometry per paraphrase.}
We plot the GAGA-PHATE latent ($\R^2$) embedding of training-prompt
activations, coloured by ground-truth weekday and panelled by
paraphrase template. The seven centroids consistently arrange around
a closed loop in latent space across paraphrases, even though
paraphrases are not aligned in any explicit cyclic-invariance loss;
the cyclic structure is recovered from PHATE diffusion distances alone.
This figure complements \cref{fig:appendix:paths2d-weekdays} by showing
that the manifold structure GAGA recovers is not an artifact of a
particular prompt phrasing.}
\label{fig:appendix:cyclic-weekdays}
\end{figure}

\begin{figure}[h]
\centering
\includegraphics[width=0.85\textwidth]{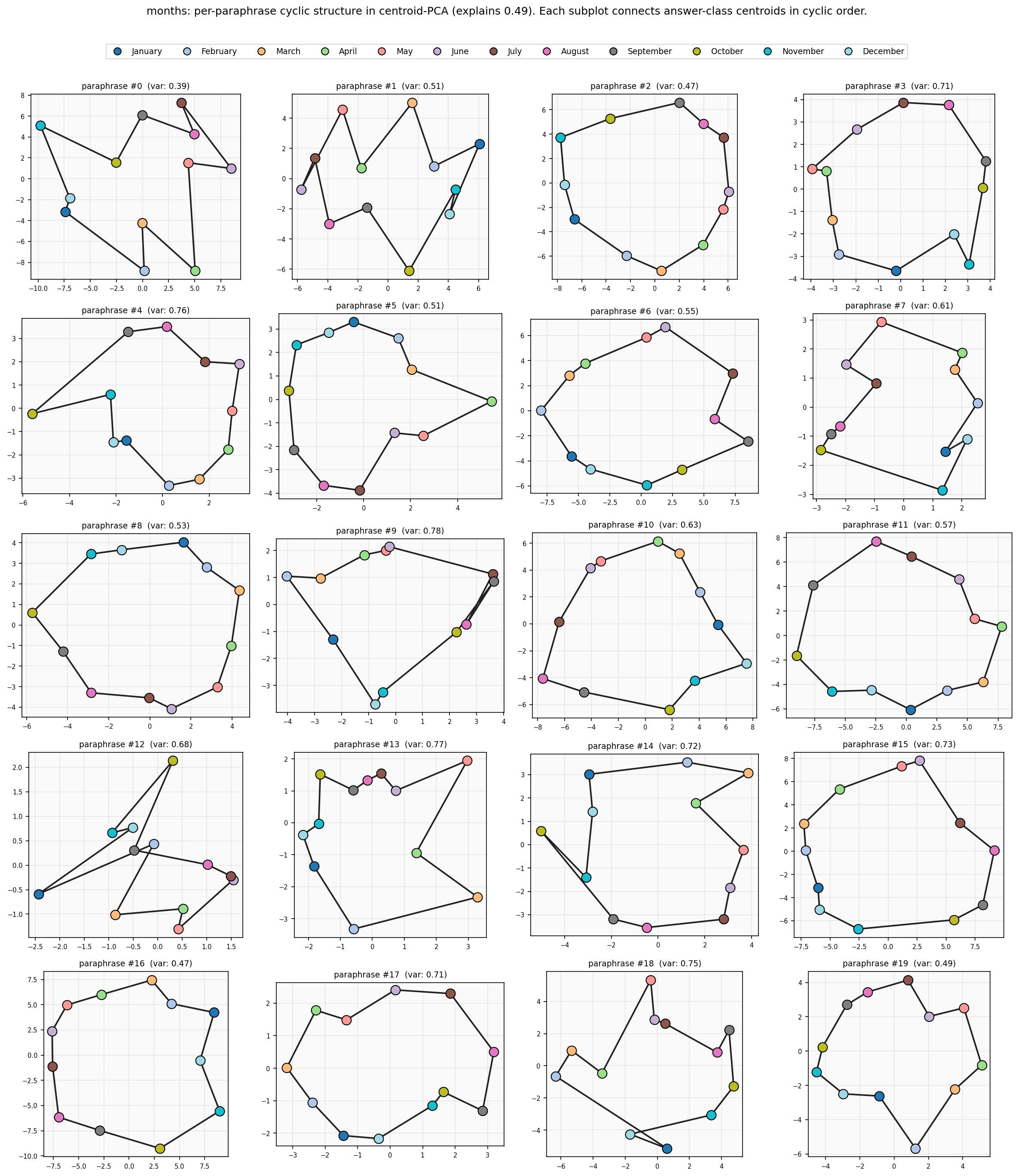}
\caption{\textbf{Months, cyclic latent geometry per paraphrase.} Same
conventions as \cref{fig:appendix:cyclic-weekdays}. Twelve months
arrange around the latent cycle with visible inter-paraphrase
consistency. Two qualitative observations: (i) cyclic spacing is not
uniform: summer months (June--August) tend to sit closer together
than winter months, suggesting the underlying activation density on
$\Mh$ is non-uniform; (ii) the cycle is consistently \emph{recovered}
but with a small per-paraphrase rotation, reflecting that PHATE's local
geometry is paraphrase-invariant up to a global symmetry.}
\label{fig:appendix:cyclic-months}
\end{figure}

\end{document}